\documentclass[11pt]{article}
\usepackage[a4paper,top=3cm,bottom=2cm,left=3cm,right=3cm,marginparwidth=1.75cm]{geometry}
\usepackage{xcolor}
\definecolor{mydarkblue}{rgb}{0.03,0.2,0.4}
\usepackage{xcolor}
\usepackage[colorlinks = true,
            linkcolor = mydarkblue,
            urlcolor  = mydarkblue,
            citecolor = mydarkblue,
            anchorcolor = mydarkblue]{hyperref}
\usepackage[round]{natbib}
\renewcommand{\bibname}{References}
\renewcommand{\bibsection}{\section*{\bibname}}
\usepackage[normalem]{ulem}
\usepackage[none]{hyphenat}
\usepackage{breakcites}
\usepackage[utf8]{inputenc} 
\usepackage[T1]{fontenc}    
\usepackage{hyperref}       
\usepackage{url}            
\usepackage{booktabs}       
\usepackage{amsfonts}       
\usepackage{nicefrac}       
\usepackage{microtype}      
\usepackage{lipsum}

\usepackage{float}
\usepackage{caption}
\usepackage{subcaption}
\usepackage{mathtools}
\usepackage{soul}
\usepackage{enumitem}
\usepackage{amssymb}
\usepackage{amsthm}
\usepackage{tikz}
\usepackage{bm}
\usetikzlibrary{arrows}
\usepackage{dsfont}
\usepackage{graphicx, graphics}
\usepackage{wrapfig}
\usepackage{thmtools,thm-restate}

\DeclareMathAlphabet{\mathbsf}{OT1}{cmss}{bx}{n}
\DeclareMathAlphabet{\mathssf}{OT1}{cmss}{m}{sl}
\usepackage{xspace}
\usepackage{amsmath,amssymb} 

\DeclareMathOperator*{\argmin}{arg\,min}
\usepackage{placeins}
\usepackage{float}
\usepackage{multirow}
\usepackage{adjustbox}
\usepackage{subcaption}
\usepackage{caption}

\newcommand{\irm}{\textsc{IRM}\xspace}
\newcommand{\irmgames}{\textsc{IRM Games}\xspace}
\newcommand{\firm}{\textsc{F-IRM Games}\xspace}
\newcommand{\virm}{\textsc{V-IRM Games}\xspace}
\newcommand{\brd}{\textsc{BRD}\xspace}
\newcommand{\fedsgd}{\textsc{FedSGD}\xspace}
\newcommand{\fedprox}{\textsc{FedProx}\xspace}
\newcommand{\fedavg}{\textsc{FedAVG}\xspace}
\newcommand{\ffictplay}{\textsc{F-FL Games (Smooth)}\xspace}
\newcommand{\vfictplay}{\textsc{V-FL Games (Smooth)}\xspace}
\newcommand{\finalalgo}{\textsc{V-FL Games (Smooth \!\!+ \!\!Fast)}\xspace}
\newcommand{\multilinecomment}[1]{}

\newcommand{\eal}{{\it et al.}}

\newcommand{\sharutalgo}{\textsc{Algorithm}\xspace}
\newcommand{\trainacc}{\textsc{Train Accuracy}\xspace}
\newcommand{\testacc}{\textsc{Test Accuracy}\xspace}
\newcommand{\flgames}{\textsc{FL Games}\xspace}
\newcommand{\coloredmnist}{\textsc{Colored MNIST}\xspace}
\newcommand{\extendedcoloredmnist}{\textsc{Extended Colored MNIST}\xspace}
\newcommand{\coloredfashionmnist}{\textsc{Colored Fashion MNIST}\xspace}

\newcommand{\cifar}{\textsc{Spurious CIFAR10}\xspace}

\usepackage{algorithm}
\usepackage[noend]{algpseudocode}
\algdef{SE}[SUBALG]{Indent}{EndIndent}{}{\algorithmicend\ }%
\algtext*{Indent}
\algtext*{EndIndent}

\usepackage{authblk}

\title{FL Games: A federated learning framework for distribution shifts}

\author[1,2,3]{Sharut Gupta}
\author[3]{Kartik Ahuja}
\author[1]{Mohammad Havaei}
\author[2]{Niladri Chatterjee}
\author[3]{Yoshua Bengio}
\affil[1]{Imagia Cybernetics, Montr\'eal, Canada}
\affil[2]{Indian Institute of Technology Delhi, New Delhi, India}
\affil[3]{Mila - Quebec AI Institute, Universit\'e de Montr\'eal\\
Quebec, Canada \\}
\affil[ ]{\newline \small{\texttt{\{sharut.gupta.mt617,niladri.chatterjee\}@maths.iitd.ac.in}} \newline \texttt{\{sharut.gupta,mohammad\}@imagia.com} \newline \texttt{\{sharut.gupta,kartik.ahuja,yoshua.bengio\}@mila.quebec}}

\begin{document}
\maketitle

\begin{abstract}
Federated learning aims to train predictive models for data that is distributed across clients, under the orchestration of a server. However, participating clients typically each hold data from a different distribution, whereby predictive models with strong in-distribution generalization can fail catastrophically on unseen domains. In this work, we argue that in order to generalize better across non-i.i.d. clients, it is imperative to only learn correlations that are stable and invariant across domains. We propose \flgames, a game-theoretic framework for federated learning for learning causal features that are invariant across clients. While training to achieve the Nash equilibrium, the traditional best response strategy suffers from high-frequency oscillations. We demonstrate that \flgames effectively resolves this challenge and exhibits smooth performance curves. Further, \flgames scales well in the number of clients, requires significantly fewer communication rounds, and is agnostic to device heterogeneity. Through empirical evaluation, we demonstrate that \flgames achieves high out-of-distribution performance on various benchmarks.\\

\end{abstract}

\section{Introduction}
With the rapid advance in technology and growing prevalence of smart devices, Federated Learning (FL) has emerged as an attractive distributed learning paradigm for machine learning models over networks of computers~\cite{kairouz2019advances,li2020federated,bonawitz2019towards}. In FL, multiple sites with local data, often known as \emph{clients}, collaborate to jointly train a shared model under the orchestration of a central hub called the \emph{server}. 

\par While FL serves as an attractive alternative to centralized training because the client data need not be sent to the server, there are several challenges associated with its optimization: 1) \emph{statistical heterogeneity} across clients; 2) \emph{massively distributed with limited communication} i.e, a large number of client devices with only a small subset of active clients at any given time \cite{mcmahan2017communication,li2020federated}. One of the most popular algorithms in this setup, Federated Averaging (\fedavg)~\cite{mcmahan2017communication}, allows multiple updates at each site  prior to communicating updates with the server. While this technique delivers huge communication gains in i.i.d. (independent and identically distributed) setting, its performance on non-i.i.d. clients is an active area of research. As shown by \cite{karimireddy2020scaffold}, client heterogeneity has direct implications on the convergence of \fedavg since it introduces a \emph{drift} in the updates of each client with respect to the server model. 
While recent works \cite{li2019feddane,karimireddy2020scaffold,yu2019linear,wang2020tackling,li2020federated,lin2020ensemble,li2019fedmd,zhu2021data} have tried to address client heterogeneity through constrained gradient optimization and knowledge distillation, most did not tackle the underlying distribution shift. These methods mostly adapt variance reduction techniques such as Stochastic Variance Reduction Gradients (SVRG)~\cite{johnson2013accelerating} to FL. 
The bias among clients is reduced by constraining the updates of each client with respect to the aggregated gradients of all other clients.
These methods can at best generalize to interpolated domains and fail to extrapolate well, i.e., generalize to new extrapolated domains~\footnote{Similar to \cite{krueger2021out}, we define interpolated domains as the domains which fall within the convex hull of training domains and extrapolated domains as those that fall outside of that convex hull.}. Further, since these works do not extract causal features, they might have poor generalization to the distribution of a test client unseen during training. 
\par According to \cite{pearl2018theoretical} and \cite{scholkopf2019causality}, learning robust predictors that are free from spurious correlations requires an altogether different approach that goes beyond standard statistical learning. In particular, they emphasize that in order to build robust systems that generalize well outside of their training environment, learning algorithms should be equipped with causal reasoning tools. Over the past year, there has been a surge in interest in bringing the machinery of causality into machine learning \cite{arjovsky2019invariant,ahuja2020invariant,scholkopf2019causality,ahuja2021linear,parascandolo2020learning,robey2021model,krueger2021out,rahimian2019distributionally}. All these approaches have focused on learning causal dependencies, which are stable across training domains and further estimate a truly invariant and causal predictor. Inspired by the invariance principle \cite{peters2016causal}, Invariant Risk Minimization (\irm) \cite{arjovsky2019invariant} introduces an alternative training objective that aims at learning representations that are invariant across domains.  Invariant Risk Minimization Games (\irmgames) reformulates the training objective proposed by \irm as an ensemble game and utilizes game-theoretic tools to learn invariant predictors (i.e., that remain accurate across domains). Formally, they establish the equivalence between the set of predictors that solve the Nash equilibrium of the ensemble game and the set of invariant predictors across the training domains.
\par Since FL typically consists of a large number of clients, it is natural for data at each client to represent different annotation tools, measuring circumstances, experimental environments and external interventions. Predictive models trained on such datasets could simply rely on spurious correlations to improve their in-distribution i.i.d. performance. Inspired by the recent progress in causal machine learning, we take a causal perspective to tackle the challenge of client heterogeneity under distribution shifts in federated learning. Specifically, we propose an algorithm for learning causal representations which are stable across clients and further enhance the generalizability of the trained model across out-of-distribution (OOD) datasets. Our algorithm draws motivation from \irmgames. We argue that, in contrast to \irm, \irmgames lends itself naturally to FL, where each client serves as a player that competes to optimize its local objective. Furthermore, the server orchestrates the optimization towards the global objective. However, despite its efficacy across multiple benchmarks, \irmgames has several key limitations that render it inapplicable in the FL setting. 
\begin{itemize}
    \item \textbf{Sequential dependency.} The underlying game theoretic algorithm in \irmgames is inherently sequential. This causes the time complexity of the algorithm to scale linearly with the number of environments.
    \item \textbf{Oscillations.} \irmgames show that large oscillations in the performance metrics emerge as the training progresses. The high frequency of these oscillations makes it difficult to define a valid stopping criterion.
    \item\textbf{Convergence speed.} The convergence speed of \irmgames has direct implications on the number of communication rounds in a FL setting. 
\end{itemize}

\noindent In this study, we introduce Federated Learning Games (\flgames) and address each of the above challenges. We summarize our main contributions below.
\begin{itemize}
    \item We propose a new framework called \flgames for learning causal representations that are invariant  across clients in a federated learning setup.
    \item Inspired from the game theory literature, we equip our algorithm to allow parallel updates across clients, further resulting in superior scalability.
    \item Using ensembles over client's historical actions, we demonstrate that \flgames appreciably smoothen the observed oscillations.
    \item By increasing the local computation at each client, we show that \flgames significantly reduces the number of communication rounds.
    \item Empirically, we show that the invariant predictors found by our approach lead to better or comparable performance than \cite{ahuja2020invariant} on several benchmarks.
\end{itemize}
\section{Related works}
\textbf{Federated learning (FL)}.
A major challenge in federated learning is data heterogeneity across clients where the local optima at each client may be far from that of the global optima in the parameter space. This causes a {\it drift} in the local updates of each client with respect to the server aggregated parameters and further results in slow and unstable convergence \cite{karimireddy2020scaffold}. Recent works have shown \fedavg to be vulnerable in such heterogeneous settings \cite{zhao2018federated}. A subset of these works that explicitly constrains gradients for bias removal are called extra gradient methods. Among these methods, \fedprox~\cite{li2020federated} imposes a quadratic penalty over the distance between server and client parameters which impedes model plasticity. Others use a form of variance reduction techniques such as SVRG~\cite{johnson2013accelerating} to regularize the client updates with respect to the gradients of other clients
~\cite{acar2021federated,li2019feddane,karimireddy2020scaffold,liang2019variance,zhang2020fedpd,konevcny2016federated}. Karimireddy~\eal~\cite{karimireddy2020scaffold} communicates to the server an additional set of variables known as control variates which contain the estimate of the update direction for both the server and the clients. Using these control variates, the drift at each client is estimated and used to correct the local updates. On the other hand, Acar~\eal~\cite{acar2021federated} estimates the drift for each client on the server and corrects the server updates. By doing so, they avoid using control variates and consume less communication bandwidth. The general strategy for variance reduction methods is to estimate client drift using gradients of other clients, and then constrain the learning objective to reduce the drift. The extra gradient methods are not explicitly optimized to discover causal features, and thus may fail when introduced with out-of-distribution examples outside the aggregated distribution of the set of clients. 
\par To date, only two scientific works \cite{francis2021towards,tenison2021gradient} have incorporated the learning of invariant predictors in order to achieve strong generalization in FL. The former adapts masked gradients \cite{parascandolo2020learning} and the latter builds on \irm to exploit invariance and improve leakage protection in FL. While \irm lacks theoretical convergence guarantees, failure modes of \cite{parascandolo2020learning} like formation of dead zones and high sensitivity to small perturbations \cite{shahtalebi2021sand}, translate to FL, rendering it unreliable.\\
\noindent \textbf{Out-of-distribution (OOD) generalization.} Generalization under distributional shift is one of the major challenges faced by machine learning systems, limiting their application in the real world. Recent research \cite{arjovsky2019invariant,ahuja2020invariant,scholkopf2019causality,ahuja2021linear,parascandolo2020learning,robey2021model,krueger2021out,rahimian2019distributionally,xie2020risk,yao2022improving,ahuja2021invariance} has tried to address this challenge by proposing alternative objectives for training mechanisms that are invariant across training environments. \irm \cite{arjovsky2019invariant} proposes finding a representation $\phi(X)$ that has good prediction abilities and also elicits an invariant predictor across environments. Works like \cite{krueger2021out,xie2020risk} propose penalty over a function of variance of training risks. \cite{ahuja2020invariant} reformulates \irm as finding the Nash equilibrium of an ensemble game played among several environments. \cite{mahajan2021domain} argues learning invariant representations for inputs derived from the same object. Recently, \cite{robey2021model} proposed Model-Based Domain Generalization, which enforces invariance to the underlying transformations of data. Another line of work \cite{rosenfeld2020risks,kamath2021does,ahuja2021invariance} has theoretically analyzed the failure modes of \irm.
\section{Background}
\subsection{Federated Averaging (\fedavg)} Federated learning methods involve a cloud server coordinating among multiple client devices to jointly train a global model without sharing data across clients. We assume that there is a cloud server that can both send and receive message from $m$ client devices. Denote $\mathcal{S}$ to be the set of client devices. Let $N_k$ denote the number of data samples at client device $k$, and $\mathcal{D}_k = \{(x_i^k, y_i^k)\}_{i=1}^{N_k}$ as it's labelled dataset. Mathematically, the objective of a FL is to minimize the approximation of global loss

\begin{equation} \label{eq:fed_opt}
 \min_w F(w) \text{ where } F(w) := \sum_{k=1}^m \frac{1}{m} \sum_{i=1}^{N_k}\frac{1}{N_k} \ell(x_i^k, y_i^k, w),
\end{equation}
where $\ell$ is the loss function and $w$ is the model parameter. One of the most popular methods in FL is Federated Averaging (\fedavg), where each client performs $E$ local-updates before communicating its weights with the server. \fedavg becomes equivalent to \fedsgd for $E=1$ wherein weights are communicated after every local update. For each client device $k$, \fedavg initializes it's corresponding device model $w^0_k$. Consequently, in round $t$, each device undergoes a local update on its dataset according to the following $$ w^{t+1}_{k} \leftarrow w^t_k - \eta^k \nabla \ell(\mathcal{B}^k_i, w^t_k), \text{  } \forall B^k_i \subseteq \mathcal{D}_k $$
where $B^k_i$ is a sampled mini-batch from $D_k$ at the $i$th step. All clients' models $\{w^{t+1}_k\}_{k \in \mathcal{S}}$ are then broadcasted to the cloud server which performs a weight average to update the global model $w^{t+1}$ as
$$ w^{t+1} \leftarrow \frac{1}{|\mathcal{S}|}\sum_{k \in \mathcal{S}} \frac{N_k}{N} w^t_k ,$$ where $N_k$ is the number of samples at client device $k$ and $N$ is the total number of samples from all clients ($N = \sum_{k=1}^m N_k$).
This aggregated global model is shared with all clients and the above process is repeated till convergence.

\subsection{Invariant Risk Minimization and Invariant Risk Minimization Games} \label{background_causality}
Consider a setup comprising datasets $\mathcal{D}_k = \{(x_i^k, y_i^k)\}_{i=1}^{N_k}$ from multiple training environments, $k \in \mathcal{E}_{tr}$ with $N_k$ being the number of samples at environment $k$. The aim of Invariant Risk Minimization (\irm) \cite{arjovsky2019invariant} is to jointly train across all these environments and learn a robust set of parameters $\theta$ that generalize well to unseen (test) environments $\mathcal{E}_{all} \supset \mathcal{E}_{tr}$. The risk of a predictor $f$ at each environment can be mathematically represented as 
$R^k(w \circ \phi) = E_{(x,y) \sim D_k} f_{\theta}((w \circ \phi); x, y)$
where $f_\theta = w \circ \phi$ is the composition of a feature extraction function $\phi: \mathcal{X} \rightarrow \mathcal{Z} \subseteq \mathbb{R}^d$ and a predictor network, $w:  \mathcal{Z} \rightarrow  \mathbb{R}^k$ where $k$ is the number of classes. \\

\noindent \textbf{Empirical Risk Minimization (ERM)} aims to minimize the average of the losses across all environments. Mathematically, the ERM objective can be formulated as $R^{\text{ERM}}(\theta) = E_{(x,y) \sim \cup_{e \in E} D_e} f_e(\theta; x, y)$. As shown in \cite{arjovsky2019invariant}, ERM fails to generalize to novel domains, which have significant distribution shift as compared to the training environments. \\

\noindent \textbf{Invariant Risk Minimization (\irm)} instead aims to capture invariant representations $\phi$ such that the optimal predictor $w$ given $\phi$ is the same across all training environments. Mathematically, they formulate the objective as a bi-level optimization problem

\begin{equation} \label{eq:irm_bilevel}
\begin{aligned}
  \min_{\phi \in \mathcal{H}_\phi, w \in \mathcal{H}_w} \sum_{k \in \mathcal{E}_{tr}} R^k(w \circ \phi) \text{ s.t.  } w \in \argmin_{w \in \mathcal{H}_w} R^k(w \circ \phi), \forall k \in \mathcal{E}_{tr} 
\end{aligned}
\end{equation}
where $\mathcal{H}_{\phi}$ , $\mathcal{H}_w$ are the hypothesis sets for feature extractors and predictors, respectively. Since each constraint calls an inner optimization routine, \irm approximates this challenging optimization problem by fixing the predictor $w$ to a scalar. \\

\noindent \textbf{Invariant Risk Minimization Games (\irmgames)} is an algorithm based on an alternate game theoretic reformulation of the optimization objective in equation \ref{eq:irm_bilevel}. It endows each environment with its own predictor $w^k \in \mathcal{H}_w$ and aims to train an ensemble model $w^{av}(z) = \frac{1}{|\mathcal{E}_{tr}|}\sum_{k=1}^{|\mathcal{E}_{tr}|} w^k(z)$ for each $z \in \mathcal{Z}$ s.t. $w^{av}$ satisfies the following optimization problem

\begin{equation} \label{eq:irmgames_bilevel}
\begin{aligned}
&\min_{w^{av},\phi\in \mathcal{H}_{\phi}} \sum_k R^k(w^{av} \circ \phi) \\
 \text {s.t. } w^k \in \argmin_{w_k' \in \mathcal{H}_w} & R^k\Bigg( \frac{1}{|\mathcal{E}_{tr}|}(w_k' + \sum_{q \in \mathcal{E}_{tr}, q \neq k} w^q)\circ \phi\Bigg), \forall k \in \mathcal{E}_{tr}
\end{aligned}
\end{equation}

The constraint in equation \ref{eq:irmgames_bilevel} is equivalent to the Nash equilibrium of a game with each environment $k$ as a player with action $w^k$, playing to maximize its utility $R^k(w^{av}, \phi)$. While there are different algorithms in the game theoretic literature to compute the Nash equilibrium, the resultant non-zero sum continuous game is solved using the best response dynamics (\brd) with clockwise updates and is referred to as \virm. 
In this training paradigm, players take turns according to a fixed cyclic order and only one player is allowed to change its action at any given time (for more details, refer to the supplement). Fixing $\phi$ to an identity map in \virm is also showed to be very effective and is called \firm.


\section{Federated Learning Games}
\subsection{Invariant risk minimization games as a natural fit for federated learning}
OOD generalization is often studied using the notion of environments. \cite{arjovsky2019invariant} formalize an environment as a data-generating distribution representing a particular location, time, context, circumstances and so forth.
Distinct environments are assumed to share some overlapping causal features. Spurious variables denote unstable features which vary across environments. A client in FL can be considered as a data-generating environment. For instance, consider a FL system with clients being hospitals, each collecting mammographic data for breast cancer detection. However, heterogeneity in acquisition systems, patient population, disease prevalence, etc can result in spurious correlations which are unique to a client. While these correlations differ across clients, features predictive of microcalcifications remain invariant.
    \par That said, \irmgames, initially developed to tackle OOD generalization, can be naturally adapted to FL. In particular, each client now serves as a player, competing to optimize its local objective, i.e. $
    \min_{w_k' \in \mathcal{H}_w}  R^k\Big( \frac{1}{|\mathcal{E}_{tr}|}(w_k' + \sum_{q \neq k} w^k)\circ \phi\Big), \forall k \in \mathcal{E}_{tr}$ (Equation \ref{eq:irmgames_bilevel}). The server acts as a mechanism designer which ensures the prediction efficacy of the models by optimizing the upper level objective of equation \ref{eq:irmgames_bilevel} i.e. $ \min_{w^{av},\phi\in \mathcal{H}_{\phi}} \sum_k R^k(w^{av} \circ \phi)$. \irmgames permit local computation at each client and yet is guaranteed to exhibit good out-of-distribution generalization behavior (See \cite{ahuja2021linear}), such guarantees do not exist for other works that explore invariance in federated learning \citep{tenison2021gradient,francis2021towards}. Further, \irmgames doesn't require additional regularization parameters for which the coefficient needs to be tuned.

\subsection{Limitations of \irmgames}
\irmgames has proved to be effective in the identification of non-spurious causal feature-target interactions across a variety of benchmarks. Although widely used in the machine learning community, there are several challenges which limit its utility in the FL domain. 
We shall elaborate on each of these limitations and propose solutions to make it feasible for its practical deployment in FL setup. \\

\noindent \textbf{Sequential dependency} \label{algo_sequential}
As discussed in Section \ref{background_causality}, \irmgames poses the \irm objective as that of finding the Nash equilibrium of an ensemble game across training environments and adopts the classic best response dynamics (\brd) algorithm to compute it. This approach is based on playing clockwise sequences wherein players take turns in a fixed cyclic order, with only one player being allowed to change its action at any given time $t$ (details in the supplement). In order to choose its optimal action for the first time steps, the last scheduled player $N$ has to wait for all the remaining players from $1,2,... N-1$ to play their strategies. This linear scaling of time complexity with the number of players poses a major challenge for adapting \irmgames to FL setup, which requires fast convergence and should exploit computational parallelism across the clients.
\par By definition, in \brd with clockwise playing sequences, the best responses of any player determine the best responses of the remaining players. Thus, in a distributed learning paradigm, the best responses of each client (player) need to be transmitted to all the other clients. This is infeasible from a practical standpoint as clients are usually based on slow or expensive connections and message transmissions can frequently get delayed or result in information loss. Hence, it is imperative to develop scalable and efficient FL algorithms which are able to learn causal features and at the same time require minimal information about the actions of other clients in the system. 
As shown on lines \ref{lst:line:parallel1} and \ref{lst:line:parallel2}(green) of Algorithm \ref{alg:main_algorithm}, we modify the classic \brd algorithm by allowing simultaneous updates at any given time $t$. We refer to this variant of \firm and \virm as \textit{parallelized} \firm and, \virm respectively.\\

\noindent \textbf{Oscillations.} As demonstrated by \cite{ahuja2020invariant}, when a neural network is trained using the \irmgames objective (equation \ref{eq:irmgames_bilevel}), the training accuracy initially stabilizes at a high value and eventually starts to oscillate. The setup over which these observations are made involves two training environments with varying degrees of spurious correlation. The environments are constructed so that the degree of correlation of color with the target label is very high. 
The explanation for these oscillations attributes the problem to the significant difference among the data of the training environments. In particular, after a few steps of training, the individual model of the environment with higher spurious correlation (say $\mathcal{E}_1$) is positively correlated with the color while the other is negatively correlated. When it is the turn of the former environment to play its optimal strategy, it tries to exploit the spurious correlation in its data and increase the weights of the neurons which are indicative of color. On the contrary, the latter tries to decrease the weights of features associated with color since the errors that backpropagate are computed over the data for which exploiting spurious correlation does not work (say $\mathcal{E}_2$). This continuous swing and sway among individual models may explain the observed oscillations.
\par Despite the promising results, with a model's performance metrics oscillating at each step, defining a reasonable stopping criterion becomes challenging. As shown in the game theoretic literature \cite{herings2017best,barron2010best,fudenberg1998theory,ge2018fictitious}, \brd can often oscillate. Computing the Nash equilibrium for general games is non-trivial and is only possible for a certain class of games (e.g., concave games) \citep{zhou2017mirror}.  Thus, rather than alleviating oscillations completely, we propose solutions to reduce them significantly to better identify valid stopping points. We propose a two-way ensemble approach wherein apart from maintaining an ensemble across clients ($w^{av}$), each client $k$ responds to the ensemble of historical models (memory) of its opponents. 
Intuitively, a moving ensemble over the historical models acts as a smoothing filter, which helps to reduce drastic variations like those discussed above. 
\par Based on the above motivation,  we reformulate the optimization objective of \irmgames (Equation \ref{eq:irmgames_bilevel}) to adapt the two-way ensemble learning mechanism (refer to line \ref{lst:line:oscillations}(red) in Algorithm \ref{alg:main_algorithm}). Formally, we maintain queues (a form of  buffer) at each client that store its historically played actions. In each iteration, a client
best responds to a uniform distribution over the past strategies of its opponents. Mathematically, the new objective can be stated as
\begin{equation} \label{eq:fictplay_bilevel}
\begin{aligned}
&\min_{w^{av},\phi\in \mathcal{H}_{\phi}} \sum_k R^k(w^{av} \circ \phi) \\
 \text {s.t. } w^k \in \argmin_{w_k' \in \mathcal{H}_w} & R^k\Bigg( \frac{1}{|\mathcal{E}_{tr}|}(w_k' + \sum_{\substack{q \in \mathcal{E}_{tr} \\ q \neq k}} w^q + \sum_{\substack{p \in \mathcal{E}_{tr} \\ p \neq k}} \frac{1}{|\mathcal{B}_p|}\sum_{j=1}^{ |\mathcal{B}_p|}w^p_j)\circ \phi\Bigg), \forall k \in \mathcal{E}_{tr}
\end{aligned}
\end{equation}
where $\mathcal{B}_q$ denotes the buffer at client $q$ and $w^q_j$ denotes the $j$th historical model of client $q$. We use the same buffer size for all clients. Moreover, as the buffer reaches its capacity, it is renewed in a first-in first-out (FIFO) manner. This variant is called \ffictplay or \vfictplay, based on the constraint on $\phi$. \\

\begin{algorithm}[htb!]
\caption[\textit{Parallelized} \textsc{FL Games (Smooth \!\!+ \!\!Fast)}\xspace]{\textit{Parallelized} \textsc{FL Games (Smooth \!\!+ \!\!Fast)}\xspace}
\label{alg:main_algorithm}
\begin{algorithmic}[1]
\State \textbf{Notations: } $\mathcal{S}$ is the set of $N$ clients;  $\mathcal{B}_k$ and $\mathcal{P}_k$ denote the buffer and information set containing copies of $\mathcal{B}_i,\forall i\neq k \in \mathcal{S}$ at client, $k$ respectively.
\State \textbf{PredictorUpdate(k):}
    \Indent 
    \State \Comment{Two-way ensemble (across time and clients) game to update predictor at each client $k$}
    \State $w_k \gets$ SGD$\Big[\ell_k \Big\{ \frac{1}{n}(w_k' + \sum_{q \in \mathcal{E}_{tr}, q \neq k} w^q +$\colorbox{red!25}{$\sum_{p \in \mathcal{E}_{tr}, p \neq k} \frac{1}{|\mathcal{B}_p|} \sum_{j=1}^{ |\mathcal{B}_p|}w^p_j$}$)\circ \phi\Big\}\Big]$ \label{lst:line:oscillations}
    \State Insert $w_k$ to $\mathcal{B}_k$, discard the oldest model in $\mathcal{B}_k$ if full
    \State return $w_k$
\EndIndent
\State \textbf{RepresentationUpdate(k):}
    \Indent 
    \State \Comment{Gradient Descent (GD) over entire local dataset at client $k$}
    \For{\colorbox{yellow!35}{every batch $b \in \mathcal{B}$}} \label{lst:line:fast}
        \State Calculate $\nabla \ell_k(w_{\text{cur}}^{\text{av}} \circ \phi_{\text{cur}};b)$ and accumulate in $\nabla\phi_k$
    \EndFor
    \State return $\nabla\phi_k$
\EndIndent
\State \textbf{Server executes:}
\Indent
    \State Initialize $w_k, \forall k \in \mathcal{S}$ and $\phi$  
    \While{round $\leq$ max-round}   
      \State \Comment{Update representation $\phi$ at even round parity}
      \If{round is even} 
        \If{Fixed-Phi}
            \State $\phi_{\text{cur}} = I$
        \EndIf
      
        \If{Variable-Phi} 
            \For{each client $k \in \mathcal{S}$ \colorbox{green!30} {in parallel}} \label{lst:line:parallel1}
                \State $\nabla \phi_k=$ RepresentationUpdate(k)
            \EndFor
            \State \Comment{Update representation $\phi$ using weighted sum of gradients across clients}
            \State $\phi_{\text{next}} = \phi_{\text{cur}} - \eta\Big(\sum_{k \in \mathcal{S}}\frac{N_k}{\sum_{j \in \mathcal{S}}N_j}  \nabla \phi_k \Big)$ 
            \State $\phi_{\text{cur}} = \phi_{\text{next}}$
        \EndIf
      
      \Else
        \For{each client $k \in \mathcal{S}$ \colorbox{green!30}{in parallel}} \label{lst:line:parallel2}
            \State $w_k$ $\gets$ PredictorUpdate(k)
        \EndFor
        \State \Comment{Client $k$ updates its information set  $\mathcal{P}_k$ by updating copies of predictors of other clients}
        \State Communicate $\forall k, \mathcal{P}_k \gets \{w_i, \forall i\neq k \in S \}$
        
      \EndIf
     \State round $\gets$ round + 1
     \State $w_{\text{curr}}^{\text{av}} = \frac{1}{N} \sum_{k \in \mathcal{S}} w_{\text{curr}}^k $
    \EndWhile
\EndIndent
\end{algorithmic}
\end{algorithm}
\noindent \textbf{Convergence speed.}
The convergence speed of \irmgames has direct implications on the number of communication rounds in the FL setup. As discussed in Section \ref{background_causality}, \cite{ahuja2020invariant} propose two variant of \irmgames: \firm and, \virm with the former being an approximation of the latter ($\phi = I$). While both approaches exhibit superior performance on a variety of benchmarks, the latter has shown its success in a variety of large scale tasks like language modeling \citep{peyrard2021invariant}. In spite of its theoretical support, \virm suffers from slower convergence due to an additional round for optimization of $\phi$. This renders it inefficient for deployment in the FL setup. To improve the efficiency of the algorithm, we propose replacing the stochastic gradient descent (SGD) over $\phi$ by a gradient descent (GD) (line \ref{lst:line:fast} (yellow) of Algorithm \ref{alg:main_algorithm}). This allows $\phi$ to be updated according to gradients accumulated across the entire dataset, as opposed to gradient step over a mini-batch. 
Intuitively, now at each gradient step, the resultant $\phi$ takes large steps in the direction of its global optimum, which in our experiments resulted in fast and stable training. This variant of \irmgames is referred to \textsc{V-FL Games (Fast)}.

\section{Experiments and Results}
\subsection{Datasets}
\cite{ahuja2020invariant} tested \irmgames over a variety of benchmarks that were synthetically constructed to incorporate color as a spurious feature. 
These included the colored digits MNIST dataset, i.e. \coloredmnist and \coloredfashionmnist, and we work with the same datasets for our experiments. Additionally, we created another benchmark, \cifar with a data-generating process resembling that of \coloredmnist. In this dataset, instead of coloring the images to establish spurious correlation, we add small black patches at various locations in the image. These locations are spuriously correlated with the label. Details on each of the datasets can be found in the supplement. For all the experiments, we report the mean performance of various baselines over 5 runs. The performance of an oracle on each of these datasets is 75\% for training and test sets.

\par  \textbf{Terminologies:} In the following analysis, the terms `Sequential' and `Parallel' denote \brd with clockwise playing sequences and simultaneous updates respectively (Lines \ref{lst:line:parallel1} and \ref{lst:line:parallel2} of Algorithm \ref{alg:main_algorithm}). Within this categorization, \firm and \virm refer to the federated adaptations of fixed and variable versions of \irmgames respectively. The approach used to smoothen out the oscillations (Line \ref{lst:line:oscillations} of Algorithm \ref{alg:main_algorithm}) is denoted by \ffictplay or \vfictplay depending on the constraint on $\phi$. The fast variant with high convergence speed is denoted \finalalgo (Line \ref{lst:line:fast} of Algorithm \ref{alg:main_algorithm}). 

\par \textbf{\coloredmnist (Table \ref{tab:coloredmnistresults})} From the table, we can observe that both \fedsgd and \fedavg are unable to generalize to the test set. Intuitively, both the approaches latch onto the spurious features to make predictions. The sequential \firm and \virm achieve 66.56 $\pm$ 1.58 and 63.78 $\pm$ 1.58 percent testing accuracy, respectively. Amongst our approaches (\ffictplay, \finalalgo, \firm (Parallel), \virm  (Parallel)), \virm  (Parallel) achieves the highest testing accuracy i.e. 68.34 $\pm$ 5.24 percent. The benefits of this approach are twofold; 1) learning a representation of truly causal features and hence good OOD generalizability, 2) scalability to large numbers of clients (detailed discussion in Section \ref{parallel_ablation}). Further, since all our proposed approaches have high testing performance, none of them relies on the spurious correlations, unlike ERM-based approaches such as \fedsgd and \fedavg.

\par \textbf{\coloredfashionmnist (Table \ref{tab:coloredfashionmnist})} Similar to the results on \coloredmnist, all the proposed algorithms achieve high testing accuracy. Among the baselines, parallelized \ffictplay achieves the highest test accuracy of 71.26 $\pm$ 4.19 percent. Apart from providing the flexibility of efficiently scaling with the number of clients, this approach also reduces the oscillations significantly (detailed discussed in Section \ref{fictplay_ablation}).

\begin{table}[!htb]
\centering
\caption{\small{\coloredmnist: Comparison of methods in terms of training and testing accuracy (mean $\pm$ std deviation).}}\label{tab:coloredmnistresults}
\begin{adjustbox}{width=0.8\textwidth}
\begin{tabular}{llll}
\toprule
\textsc{Type} &\sharutalgo & \trainacc & \testacc\\
\midrule
 &\fedsgd & 84.88 $\pm$ 0.16 &  10.45 $\pm$ 0.60\\
 &\fedavg & 84.45 $\pm$ 2.69 &  12.52 $\pm$	4.34\\
\midrule
\multirow{4}{*}{\rotatebox[origin=c]{90}{Sequential}} & \firm & 55.76 $\pm$ 2.03 & 66.56 $\pm$ 1.58\\
 & \virm & 56.40 $\pm$  0.03 & 63.78 $\pm$ 1.58 \\
 &\ffictplay & 62.83 $\pm$ 5.06 & 66.83 $\pm$ 1.83 \\
 &\finalalgo & 61.03 $\pm$ 3.11 &  65.81 $\pm$ 3.28 \\
 \midrule
 \multirow{4}{*}{\rotatebox[origin=c]{90}{Parallel}} & \firm & 58.03 $\pm$ 6.22 & 67.14 $\pm$ 2.95\\
 & \virm & 52.89 $\pm$ 8.03 & 68.34 $\pm$ 5.24 \\
 &\ffictplay & 61.07 $\pm$ 1.71 & 67.21 $\pm$  2.98\\
 &\finalalgo & 63.11 $\pm$  3.02 & 65.73 $\pm$ 1.53 \\
\bottomrule
\end{tabular}
\end{adjustbox}
\end{table}

\begin{table}[!htb]
\centering
\caption{\small{\coloredfashionmnist: Comparison of methods in terms of training and testing accuracy (mean $\pm$ std deviation).}}\label{tab:coloredfashionmnist}
\begin{adjustbox}{width=0.8\textwidth}
\begin{tabular}{llll}
\toprule
\textsc{Type} &\sharutalgo & \trainacc & \testacc\\
\midrule
 &\fedsgd & 83.49 $\pm$ 1.22 & 20.13 $\pm$ 8.06 \\
 &\fedavg & 86.225 $\pm$ 0.63 &	13.33	$\pm$ 2.07 \\
\midrule
\multirow{4}{*}{\rotatebox[origin=c]{90}{Sequential}} & \firm & 75.13 $\pm$ 1.38 &  68.40 $\pm$ 1.83\\
 & \virm & 69.90 $\pm$ 4.56 & 69.90 $\pm$ 1.31 \\
 &\ffictplay & 75.18 $\pm$ 0.37 & 71.81 $\pm$ 1.60\\
 &\finalalgo & 75.10  $\pm$ 0.48 & 69.85 $\pm$ 1.22\\
 \midrule
 \multirow{4}{*}{\rotatebox[origin=c]{90}{Parallel}} & \firm & 71.71 $\pm$ 8.23 &  69.73 $\pm$ 2.12\\
 & \virm & 66.33 $\pm$ 9.39 & 69.85 $\pm$ 3.42 \\
 &\ffictplay &72.81 $\pm$ 4.51 &  71.36 $\pm$ 4.19\\
 &\finalalgo &  71.89 $\pm$ 5.58 & 69.41 $\pm$ 5.49 \\
\bottomrule
\end{tabular}
\end{adjustbox}
\end{table}


\par \textbf{\cifar (Table \ref{tab:cifar})} Both ERM-based approaches, \fedsgd and \fedavg achieve poor testing accuracy. However, \firm (Parallel) achieves the highest testing accuracy of 52.07 $\pm$ 1.60 percent. \\

\begin{table}[!htb]
\centering
\caption{\small{\cifar: Comparison of methods in terms of training and testing accuracy (mean $\pm$ std deviation).}}\label{tab:cifar}
\begin{adjustbox}{width=0.8\textwidth}
\begin{tabular}{{llll}}
\toprule
\textsc{Type} &\sharutalgo & \trainacc & \testacc\\
\midrule
 &\fedsgd & 84.79 $\pm$ 0.17 & 12.57 $\pm$ 0.55\\
 &\fedavg & 85.41 $\pm$ 1.45 & 13.11 $\pm$ 1.82\\
\midrule
\multirow{4}{*}{\rotatebox[origin=c]{90}{Sequential}} & \firm & 50.36 $\pm$ 2.78 & 47.36 $\pm$ 4.33 \\
 & \virm & 61.72 $\pm$ 7.39 & 46.07 $\pm$ 6.01 \\
 &\ffictplay & 64.02 $\pm$ 2.08& 45.54 $\pm$ 1.04 \\
 &\finalalgo & 50.37 $\pm$ 4.97 & 50.94 $\pm$ 3.28\\
 \midrule
 \multirow{4}{*}{\rotatebox[origin=c]{90}{Parallel}} & \firm & 55.06 $\pm$ 2.04 & 52.07 $\pm$ 1.60\\
 & \virm & 50.41 $\pm$ 3.31&  50.43 $\pm$ 3.04\\
 &\ffictplay & 56.98 $\pm$ 4.09 & 49.56 $\pm$ 1.36\\
 &\finalalgo & 45.83 $\pm$ 2.44 & 49.89 $\pm$5.66 \\
\bottomrule
\end{tabular}
\end{adjustbox}
\end{table}

\noindent In all the above experiments, our main algorithm i.e. \textit{parallelized \finalalgo} is able to achieve high testing accuracy which is comparable to the maximum performance. Hence, the benefits provided by this approach in terms of 1) robust predictions; 2) scalability; 3) fewer oscillations and 4) fast convergence are not at the cost of performance. While the former is demonstrated by Tables \ref{tab:coloredmnistresults}, \ref{tab:coloredfashionmnist} and \ref{tab:cifar}, the latter three are detailed in the following section.

\subsection{Ablation Analysis} \label{ablation}
In this section, we analyze the effect of each of our algorithmic modifications using illustrative figures on the \coloredmnist dataset. The results on other datasets are similar and are provided in the supplement. 

\subsubsection{Effect of Simultaneous \brd} \label{parallel_ablation}
\begin{figure}[htb!]
\centering
  \subfloat[]{\includegraphics[width=0.45\textwidth]{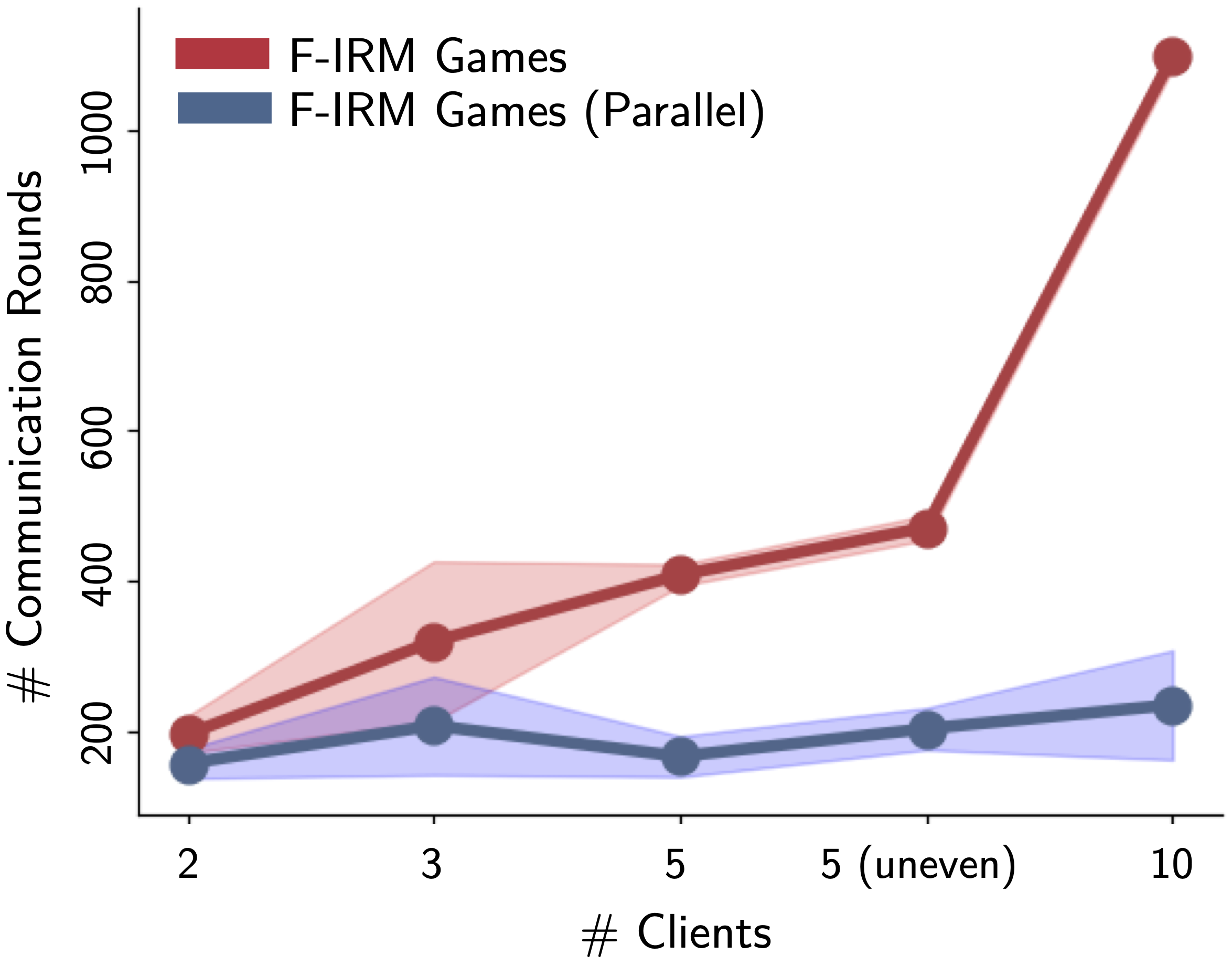}\label{fig:sequential_vs_parallel_ablation}}
  \hfill
  \subfloat[]{\begin{adjustbox}{width=0.50\textwidth}
\begin{tabular}{llll}
\toprule
 Type & \# clients & \trainacc & \testacc  \\
\midrule
\multirow{5}{*}{\rotatebox[origin=c]{90}{Sequential}} &  2	&	53.81	$\pm$	4.14	&	65.68	$\pm$	1.89	\\
	&	3	&	54.9	$\pm$	4.37	&	66.33	$\pm$	1.24	\\
	&	5	&	57.2	$\pm$	2.07	&	66.53	$\pm$	0.55	\\
	&	5 (uneven)	&	58.09	$\pm$	2.21	&	65.3	$\pm$	2.08	\\
	&	10	&	59.39	$\pm$	1.41	&	66.57	$\pm$	1.02	\\
\midrule
\multirow{5}{*}{\rotatebox[origin=c]{90}{Parallel}}   & 2	&	57.95	$\pm$	3.46	&	66.57	$\pm$	2.99	\\
	&	3	&	59.67	$\pm$	5.46	&	65.35	$\pm$	3.73	\\
	&	5	&	61.82	$\pm$	4.29	&	65.53	$\pm$	3.85	\\
	&	5 (uneven)	&	56.96	$\pm$	5.61	&	66.15	$\pm$	3.95	\\
	&	10	&	55.24	$\pm$	2.88	&	67.49	$\pm$	3.02	\\
\bottomrule
\end{tabular}
\end{adjustbox}}
\caption{\small{\coloredmnist: (a) Number of communication rounds required to achieve to achieve Nash equilibrium versus the number of clients in the FL setup; (b) Comparison of \firm and \firm (Parallel) with increasing clients in terms of training and testing accuracy (mean $\pm$ std deviation).}}
\label{fig:parallel_vs_sequential_ablation}
\end{figure}

We examine the effect of replacing the classic best response dynamics, following \cite{ahuja2020invariant}, with the simultaneous best response dynamics. For these experiments, we use a more practical environment: (a) more clients are involved, and (b) each client has fewer data. Similar to \cite{choe2020empirical}, we extended the \coloredmnist dataset by varying the number of clients between 2 and 10. For each setup, we vary the degree of spurious correlation between 70\% and 90\% for training clients while introducing a mere 10\% spurious correlation in the test set.  A more detailed discussion of the dataset is provided in the supplement. For \firm, it can be observed from Figure \ref{fig:parallel_vs_sequential_ablation}(a), as the number of clients in the FL system increases, there is a sharp increase in the number of communication rounds required to reach an equilibrium. However, the same doesn't hold true for \textit{parallelized} \firm. Further,  \textit{parallelized} \firm is able to reach a comparable or higher test accuracy as compared to \firm with significantly lower communication rounds (refer to Table \ref{fig:parallel_vs_sequential_ablation}(b)).

\subsubsection{Effect of a memory ensemble} \label{fictplay_ablation}
\begin{figure}[htb!]
\centering
  \subfloat[]{\includegraphics[width=0.5\textwidth]{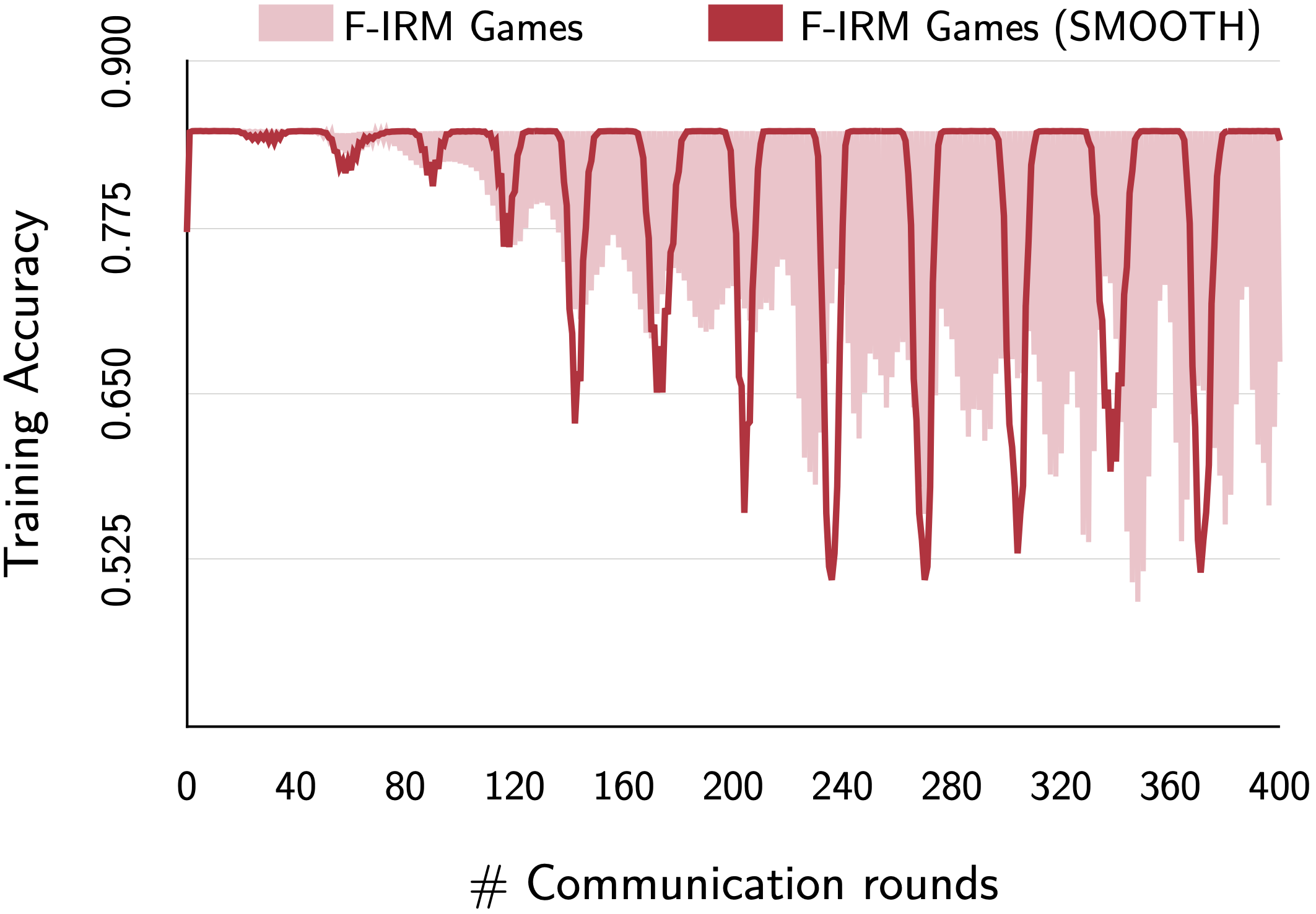}\label{fig:batchnorm_figure}}
  \hfill
  \subfloat[]{\includegraphics[width=0.5\textwidth]{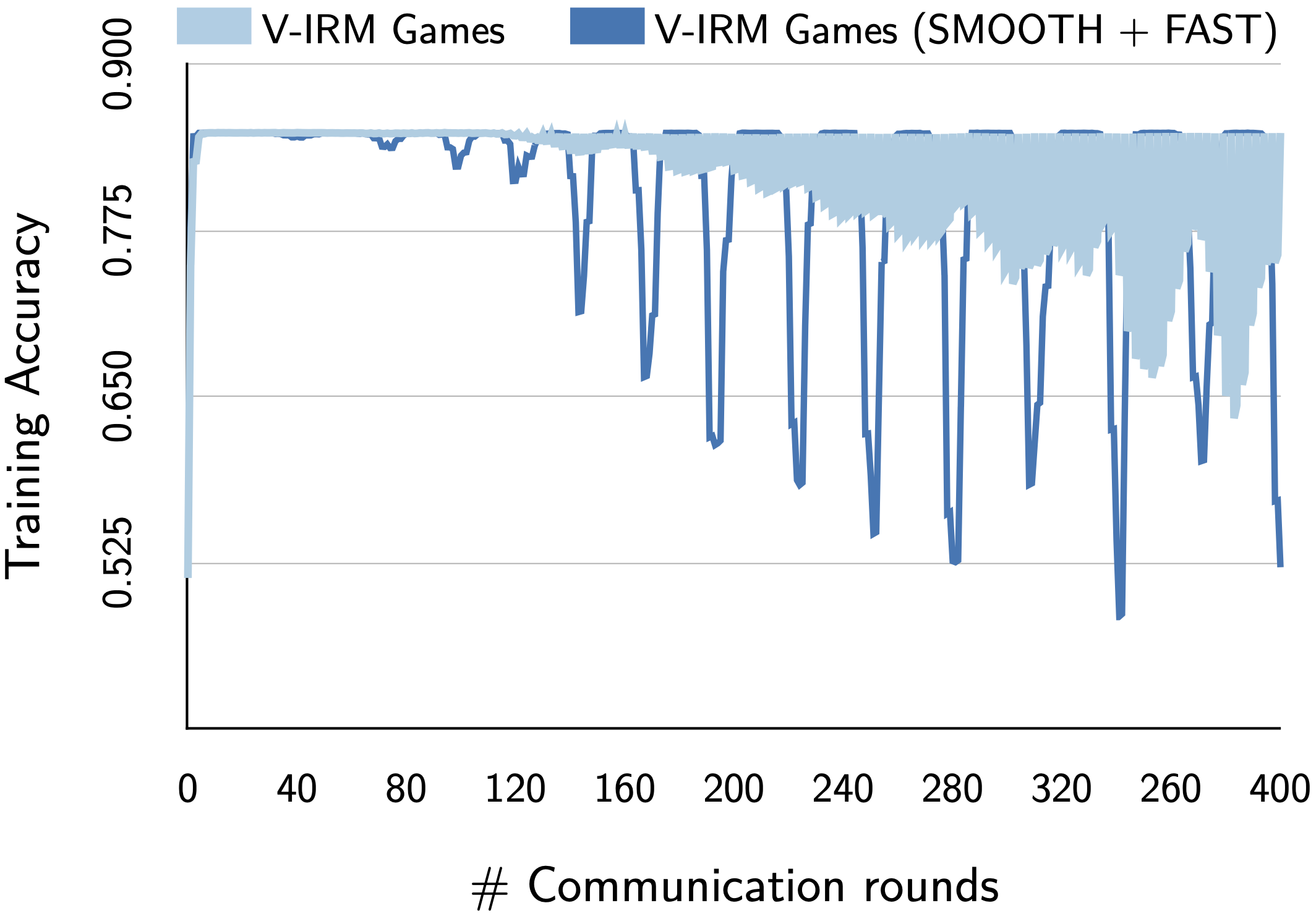}\label{fig:loss_landscape}}
\caption{\small{\coloredmnist: Training accuracy of (a)\firm and \ffictplay for a buffer size of 5; (b) \virm and \finalalgo with buffer size as 5 versus the number of communication rounds.}}
\label{fig:fast_illustration_ablation}
\end{figure}

As shown in Figure \ref{fig:fast_illustration_ablation}(a), compared to \firm, \ffictplay reduces the oscillations significantly. In particular, while in the former, performance metrics oscillate at each step, the oscillations in the latter are observed only after an interval of roughly 50 rounds. Further, \ffictplay seems to envelop the performance curves of \irmgames. As a result, apart from reducing the frequency of oscillations, \ffictplay also achieves higher testing accuracy. This implies that it does not rely on the spurious features to make predictions. Similar performance evolution curves are also observed for \textit{parallelized} \ffictplay with an added benefit of faster convergence as compared to \textit{parallelized} \firm.


\subsubsection{Effect of using Gradient Descent (GD) for $\phi$}
Communication costs are the principal constraints in FL setup. Edge devices like mobile phones and sensor are bandwidth constrained and require more power for transmission and reception as compared to remote computation. As observed from Figure  \ref{fig:fast_illustration_ablation}(b), \finalalgo is able to achieve significantly higher testing accuracy in fewer communication rounds as compared to \virm. 

\subsubsection{Effect of exact best response}
\begin{figure}[htb!]
\centering
  \subfloat[]{\includegraphics[width=0.49\textwidth]{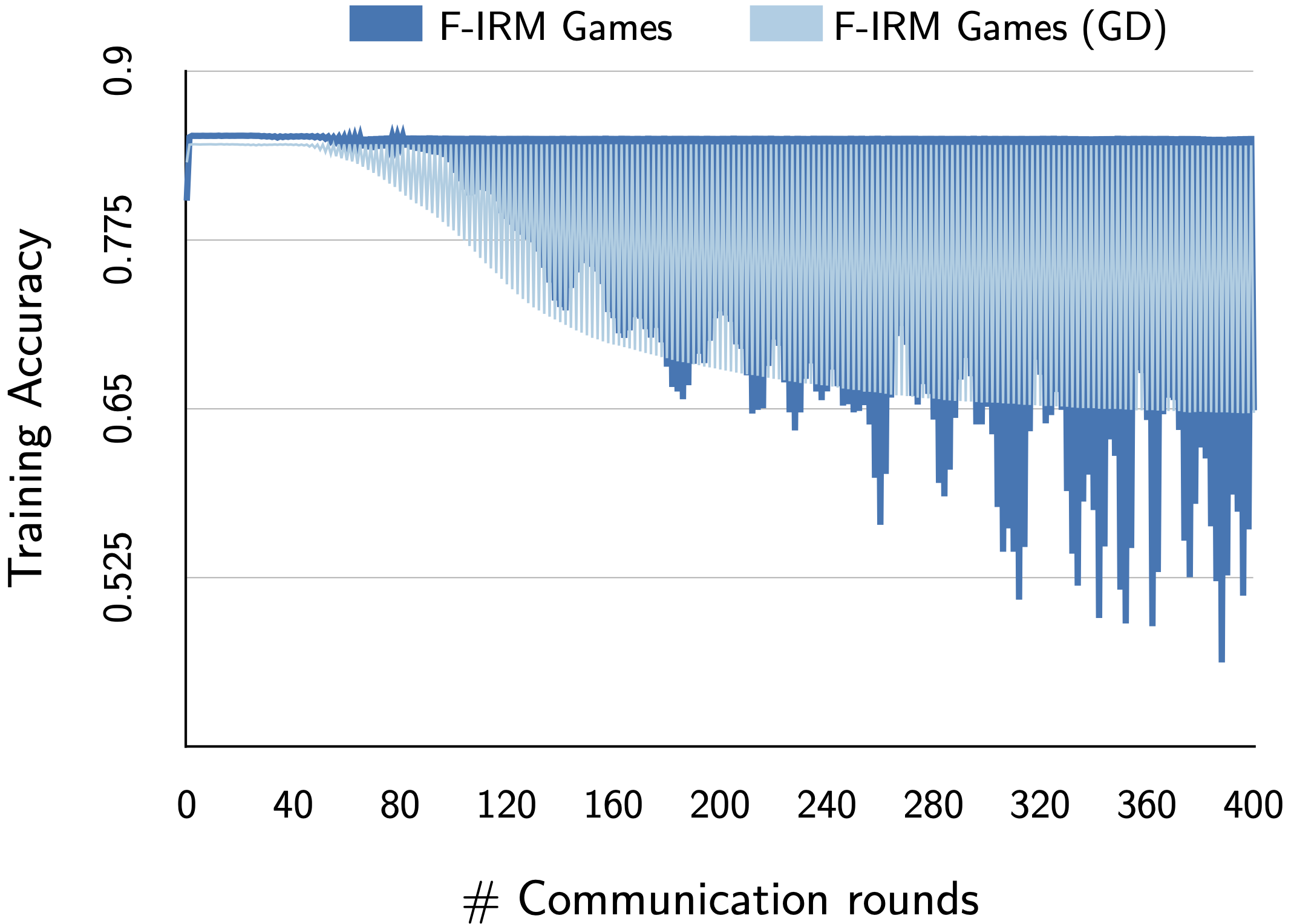}\label{fig:more_local_compute}}
  \hfill
  \subfloat[]{\begin{adjustbox}{width=0.49\textwidth}
\begin{tabular}{{lll}}
\toprule
Local steps (in \%) & \trainacc & \testacc \\
\midrule
1.71\%	&	53.04	$\pm$	1.85	&	65.05	$\pm$	1.60	\\
4.27\%	&	55.12	$\pm$	4.76	&	61.37	$\pm$	5.02	\\
6.84\%	&	53.46	$\pm$	3.66	&	61.64	$\pm$	3.67	\\
8.55\%	&	54.13	$\pm$	5.2	&	61.07	$\pm$	5.89	\\
19.66\%	&	54.38	$\pm$	6.93	&	59.17	$\pm$	7.37	\\
29.91\%	&	53.86	$\pm$	8.14	&	58.40	$\pm$	7.19	\\
49.57\%	&	53.59	$\pm$	8.56	&	59.22	$\pm$	6.05	\\
70.09\%	&	54.25	$\pm$	8.21	&	58.66	$\pm$	7.09	\\
80.34\%	&	53.14	$\pm$	9.86	&	57.93	$\pm$	6.23	\\
89.74\%	&	54.61	$\pm$	7.32	&	57.76	$\pm$	6.54	\\
100.00\%	&	54.56	$\pm$	6.56	&	57.02	$\pm$	5.17	\\
\bottomrule
\end{tabular}
\end{adjustbox}}
\caption{\small{\coloredmnist: (a) Effect on Training accuracy of doing a gradient descent on each client for updating the predictor versus the standard training paradigm i.e. \firm ;(b) Impact of increasing the number of local steps (in \% of the maximum number of steps} for updating the predictor on the training and testing accuracy (mean $\pm$ std deviation). When the number of local steps (in \%) reaches 100\%, it is equivalent to a gradient descent as shown in (a)}
\label{fig:more_local_computation_ablation}
\end{figure}

\fedavg provides the flexibility to train communication efficient and high-quality models by allowing more local computation at each client. This is particularly detrimental in scenarios with poor network connectivity, wherein communicating at every short time span is infeasible. Inspired by \fedavg, we study the effect of increasing the amount of local computation at each client. Specifically, in \firm, each client updates its predictor based on a step of stochastic gradient descent over its mini-batch. We modify this setup by allowing each client to run a few steps of stochastic gradient descent locally. When the number of local steps at each client reaches is maximum (training data size/ mini-batch size)), the scenario becomes equivalent to a gradient descent (GD) over the training data. From Table \ref{fig:more_local_computation_ablation}(b), it is evident that as the number of local steps increase i.e. each client \textbf{exactly} best responds to its opponents, the testing accuracy at equilibrium starts to decrease. 
When the local computation reaches 100\%, i.e. each client updates its local predictor based on a GD over its data, \firm exhibits reassuring convergence (as shown in Figure \ref{fig:more_local_computation_ablation}(a)). \irmgames is guaranteed to exhibit convergence and good out-of-distribution generalization behavior \citep{ahuja2021linear} despite increasing local computations. Although the testing accuracy at convergence is lower compared to the standard setup, this approach opens avenues for practical deployment of the approach in FL.  

\section{Conclusion}
In this work, we develop a novel framework based on the Best Response Dynamics (\brd) training paradigm to learn invariant predictors across clients in Federated learning (FL). Inspired by \cite{ahuja2020invariant}, the proposed method called Federated Learning Games (\flgames) learns causal representations which have good out-of-distribution generalization on new training clients or for test clients unseen during training. We investigate the high frequency oscillations observed using \brd and equip our algorithm with a memory of historical actions. This results in smoother evolution of performance metrics, with significantly lower oscillations. \flgames exhibits high communication efficiency as it allows parallel computation, scales well in the number of clients and results in faster convergence. Given the impact of FL in medical imaging, we plan to test our framework over medical benchmarks. Future directions include theoretically analyzing the smoothed best response dynamics, as it might have potential implications for other game-theoretic based machine learning frameworks. 


\newpage
\bibliography{refs}
\bibliographystyle{iclr2022_conference}
\newpage

\section{Supplementary}
\subsection{Game Theory Concepts} \label{background_gametheory}

We define some basic game theory notations that will be used later. \\
Let $\Gamma = (N, \{S_k\}_{k \in N}, \{u_k\}_{k \in N})$ be the tuple representing a normal form game, where $N$ denotes the finite set of players.  For each player $k$, $S_k = \{s^k_0, s^k_1,...s^k_m\}$ denotes the pure strategy space with $m$ strategies and $u_k(s_k, s_{-k})$ denotes the  payoff function of player $k$ corresponding to strategy $s_k$. Here, an environment for player $k$ is $s_{-k}$, a set containing strategies taken by all players but $k$ and $S_{-k}$ denotes the space of strategies of the opponent players to $k$. $S = \prod_{i \in N} S_i$ denotes the joint strategy set of all players. A game $\Gamma$ is said to be finite if $S$ is finite and is continuous if $S$ is uncountably infinite.
\par While a pure strategy defines a specific action to be followed at any time instance, a mixed strategy of player $k$, $\sigma_k = \{p_k(s^k_0), p_k(s^k_1), ... p_k(s^k_m)\}$ is a probability distribution over a set of pure strategies, where $\sum_{j=1}^m p_k(s^k_j)=1$. The expected utility of a mixed strategy $u_k(\sigma_k, \sigma_{-k})$ for player $k$ is the expected value of the corresponding pure strategy payoff i.e.
$$ \mathbb{E}(u_k(\sigma_k, \sigma_{-k})) = \sum_{s_k \in S_k} \sum_{s_{-k} \in S_{-k}} u_k(s_k, s_{-k})p_k(s_k) p_{-k}(s_{-k}), \forall \sigma_k \in \Tilde{S}_k$$
where $\Tilde{S}_k$ corresponds to the mixed strategy space of player $k$.

\noindent \textbf{Best response (BR).} A mixed strategy $\sigma^*_k$ for player $k$ is said to be a best response to it's opponent strategies $\sigma_{-k}$ if 
$$ \mathbb{E}(u_k(\sigma^*_k, \sigma_{-k})) \geq \mathbb{E}(u_k(\sigma_k, \sigma_{-k})), \forall \sigma_k \in \Tilde{S}_k.$$

\noindent \textbf{Nash equilibrium.} A mixed strategy profile $\sigma^* = \{\sigma^*_1,\sigma^*_2,...\sigma^*_N\}$ is a Nash equilibrium if for all players $k$, $\sigma^*_k$ is the best response to the strategies played by it's opponent players i.e $\sigma^*_{-k}$.\\

\noindent \textbf{Best response dynamics (\brd).}  \brd is an iterative algorithm in which at each time step, a player myopically plays strategies that are best responses to the most recent known strategies played by it's opponents previously. Based on the playing sequence across layers, \brd can be classified into three broad categories: \brd with clockwise sequences, \brd with simultaneous updating and \brd with random sequences. For this study, we focus only on the first two playing schedules. Let the function $\text{seq}: \mathbb{N} \rightarrow \mathcal{P}[N]$ denote a playing sequence which determines the set of players whose turn it is to play at each time period $t \in \mathbb{N}$. Here, $\mathcal{P}[N]$ denotes the power set of $\{1,2,...N\}$ players and $\mathbb{N}$ be the set of natural numbers $\{1,2,...\}$. By \brd, at each time step $t$, $\forall i \in \text{seq}(t)$, action taken by player $i$ i.e. $a_i^t$ is the best response to it's current environment i.e $a_{-i}^t$. 
\begin{itemize}
    \item \brd with clockwise sequences: In this playing sequence, players take turns according to a fixed cyclic order and only one player is allowed to change it's action at any given time $t$. Specifically, the playing sequence is defined by $\text{seq}(t) = 1 + (t-1) \mod n$. Since only a single player is allowed to play at any given time $t$, $a_{\text{-seq}(t)}^t = a_{\text{-seq}(t)}^{t-1}$. 
    \item \brd with simultaneous updating: In this playing sequence, $\text{seq}(t)$ chooses a non empty subset of players to participate in round $t$. However, for each player $i \in \text{seq}(t)$, the optimal action chosen $a_i^t$ depends on the knowledge of the latest strategy of it's opponents.
\end{itemize}

\subsection{Datasets}
The MNIST dataset consists of handwritten digits, with a total of 60,000 images in the training set and 10,000 images in the test set \cite{deng2012mnist}. These images are black and white in colour and form a subset of the larger collection of digits called NIST. Each digit in the dataset is normalized in size to centre fit in the fixed size image of size 28$\times$28. It is then anti-aliased to introduce appropriate gray-scale levels. 

\subsubsection{\coloredmnist}
\begin{figure}[htb!]
    \centering
\includegraphics[width=0.55\linewidth]{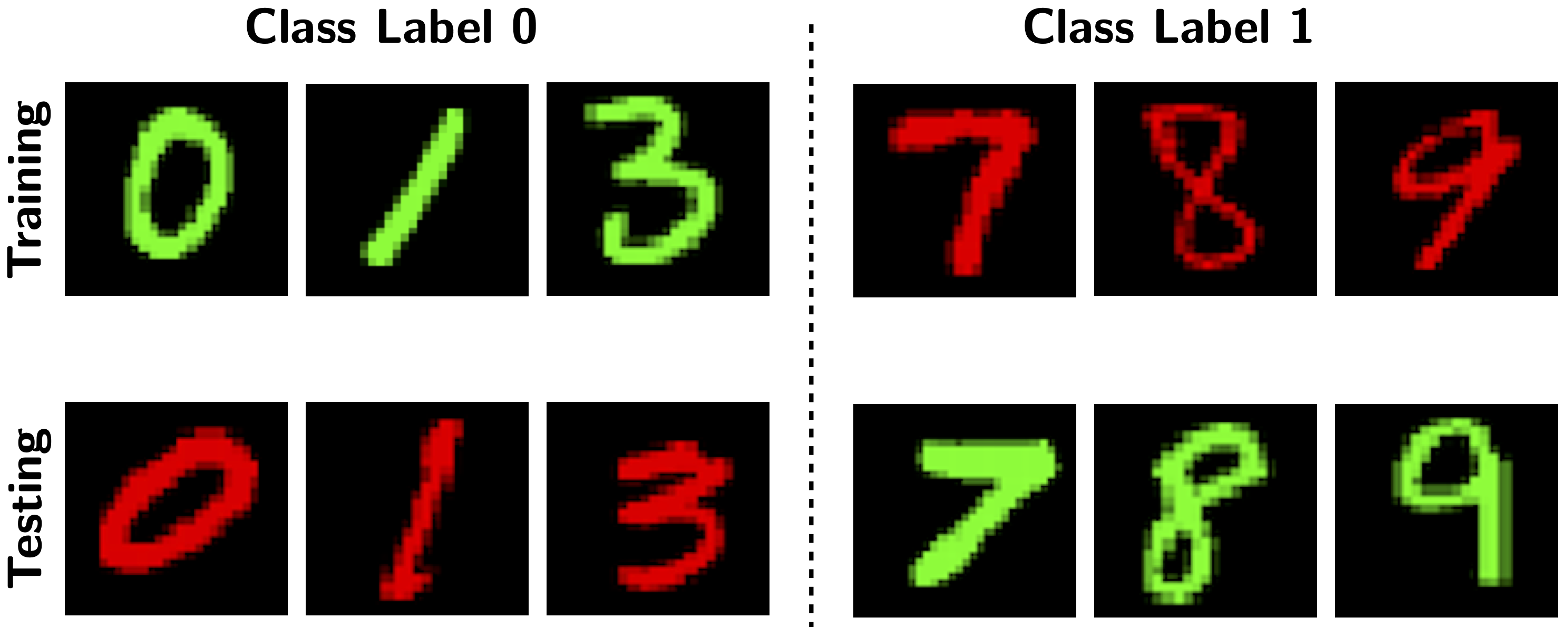}
    \caption{\small{\coloredmnist: Illustration of samples containing high spurious correlation between label and color during training. While testing, this correlation is significantly reduced as label 0 is highly correlated with the color mapped to label 1 and vice-versa.}}
    \label{fig:colored_mnist_dataset}
\end{figure}

We modify the MNIST dataset in the exact same manner as in \cite{arjovsky2019invariant}. Specifically, \cite{arjovsky2019invariant} creates the dataset in a way that it contains both the invariant and spurious features according to different causal graphs. Spurious features are introduced using colors. Digits less than 5 (excluding 5) are attributed with label 0 and the rest with label 1. The dataset is divided across three clients, out of which two serve as training and one as testing. The 60,000 images from MNIST train set are divided equally amongst the two training clients i.e. each consists of 30,000 samples. The testing set contains the 10,000 images from the MNIST test set. Preliminary noise is added to the label to reduce the invariant correlation. Specifically, the initial label ($\Tilde{y}$) of each image is flipped with a probability $\delta_k$ to construct the final label $y$. The final label $y$ of each image is further flipped with a probability $p_k$ to construct it's color code ($z$). In particular, the image is colored red, if $z=1$ and green if $z=0$.
The flipping probability which defines the color coding of an image, $p_k$ is 0.2 for client 1, 0.1 for client 2 and 0.9 for the test client. The probability $\delta_k$ is fixed to 0.25 for all clients $k$. The above choice is defined in a way that the mean degree of label-color (spurious) correlation ($1-p_k, \forall k$) is more than the average degree of invariant correlation ($1-\delta_k, \forall k$). A sample batch of images elucidating the above construction is shown in Figure \ref{fig:colored_mnist_dataset}.

\subsubsection{\coloredfashionmnist}
\begin{figure}[htb!]
    \centering
\includegraphics[width=0.55\linewidth]{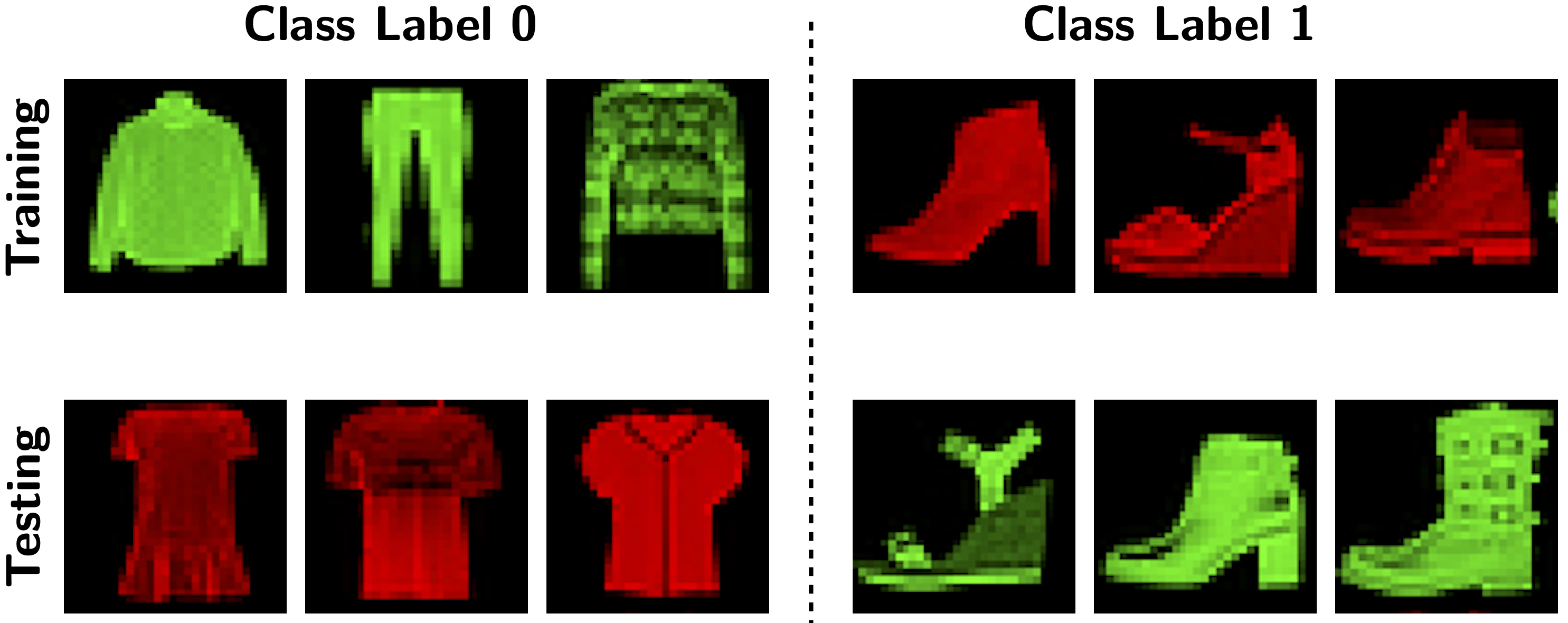}
    \caption{\small{\coloredmnist: Illustration of samples containing high spurious correlation between label and color during training. While testing, this correlation is significantly reduced as label 0 is highly correlated with the color mapped to label 1 and vice-versa.}}
    \label{fig:colored_fashion_mnist_dataset}
\end{figure}

We use the exact same environment for creating \coloredfashionmnist as in \cite{ahuja2020invariant}. The data generating process of \coloredfashionmnist is motivated from that of \coloredmnist in a way that it possesses spurious correlations between the label and the colour. Fashion MNIST consists of images from a variety of sub-categories under the two broad umbrellas of clothing and footwear. Clothing items include categories like: ``t-shirt", ``trouser", ``pullover", ``dress", ``shirt" and ``coat" while the footwear category includes ``sandal", ``sneaker", ``bag" and ``ankle boots". Similar to \coloredmnist, the train dataset is equally split across two clients (30,000 images each) and the entire test set is attributed to the test client. Preliminary labels for binary classification are constructed such  $\Tilde{y} = 0$ for “t-shirt”, “trouser”, “pullover”, “dress”, “coat”, “shirt” and $\Tilde{y} = 1$: “sandle”, “sneaker” and  “ankle boots”. Next, we add noise to the preliminary label by flipping $\Tilde{y}$ with a probability $\delta_k = 0.25, \forall k$ to construct the final label $y$. We next flip the final label with a probability $p_k$ to designate a color ($z$), with $p_1 = 0.2$ for the first client, $p_2 = 0.1$ for the second client and $p_3 = 0.9$ for the test client. The image is colored red, if $z=1$ and green if $z=0$. A sample batch of images elucidating the above construction is shown in Figure \ref{fig:colored_fashion_mnist_dataset}.

\subsubsection{\cifar} 
\begin{figure}[htb!]
    \centering
\includegraphics[width=0.55\linewidth]{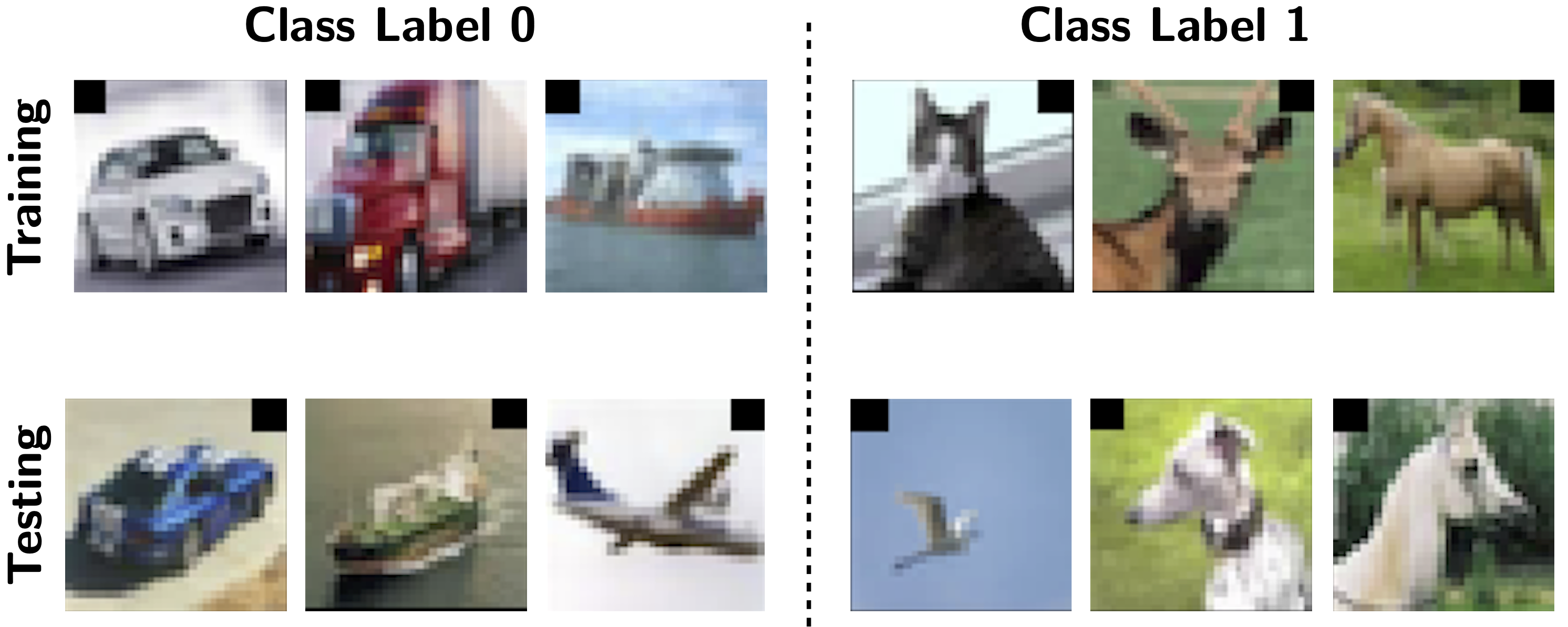}
    \caption{\small{\cifar: Illustration of samples containing high spurious correlation between labels and spatial location of the 5$\times$5 black patch during training. While testing, this correlation is significantly reduced as label 0 is now highly correlated with the spatial location corresponding to label 1 and vice-versa.}}
    \label{fig:spurious_cifar10_dataset}
\end{figure}

In this setup, we modify the CIFAR-10 dataset similar to the \coloredmnist dataset. Instead of coloring the images, we use a different mechanism based on the spatial location of a synthetic feature to generate spurious features. CIFAR10 dataset consists of 60,000 images from 10 classes including ``airplane", ``automobile", ``bird", ``cat", ``deer", ``dog", ``frog", ``house", ``ship", ``truck". The original dataset is relabelled to create a binary classification task between motor and non-motor objects. All images corresponding to the label ``frog" are discarded to ensure a similar samples count for the two classes. Similar to \coloredmnist, the train dataset is equally split across two clients and the entire test set is attributed to the test client. Preliminary labels for binary classification are constructed such  $\Tilde{y} = 0$ for “airplane”, “automobile”, “ship”, “truck” and $\Tilde{y} = 1$: “bird”, “cat”, ``deer",``dog" and  “horse”. Next, we add noise to the preliminary label by flipping $\Tilde{y}$ with a probability $\delta_k = 0.25, \forall k$ to construct the final label $y$. We next flip the final label with a probability $p_k$ to designate a positional index ($z$), with $p_1 = 0.2$ for the first client, $p_2 = 0.1$ for the second client and $p_3 = 0.9$ for the test client. This index defines the spatial location of a 5$\times$5 black patch in the image. An index value of 0 ($z=0$) specifies the patch at the top left corner of the image, while an index value of 1 ($z=1$) corresponds to a black patch over the top-right corner of the image. Based on the choice of flipping probabilities, images in the training set are spuriously correlated to the position of the patch in the image. Such correlation does not exist while testing. A sample batch of images elucidating the above construction is shown in Figure \ref{fig:spurious_cifar10_dataset}.

\subsubsection{Extended Datasets} \label{multiclassdatasets}
We extend all the three datasets \coloredmnist, \coloredfashionmnist and \cifar as in \cite{choe2020empirical} to robustly test our approach across multiple clients. Analogous to \coloredmnist with two training environments ($N=2$), we extend the datasets to incorporate $N=2,3,5$ and $10$ training clients. In particular, we attribute each client with a unique flipping probability $p_k$. For each value of $N$, the maximum value of $p_k, \forall k$ is 0.3, while the minimum value is 0.1. The values of $p_k$ for each client are spaced evenly between this range. For example, for the case of $N=5$ clients, the flipping probabilities of clients are $p_1 = 0.3$, $p_2 = 0.2$, and $p_3 = 0.1$. The flipping probability $\delta_k$ which decides the final label $y$ is fixed to 0.25 for all clients. The maximum and the minimum values of $p_k$ are chosen in a way that the average spurious correlation which is 0.8 is more than the invariant correlation i.e. 0.75.
\par
\begin{figure}[htb!]
    \centering
\includegraphics[width=1.0\linewidth]{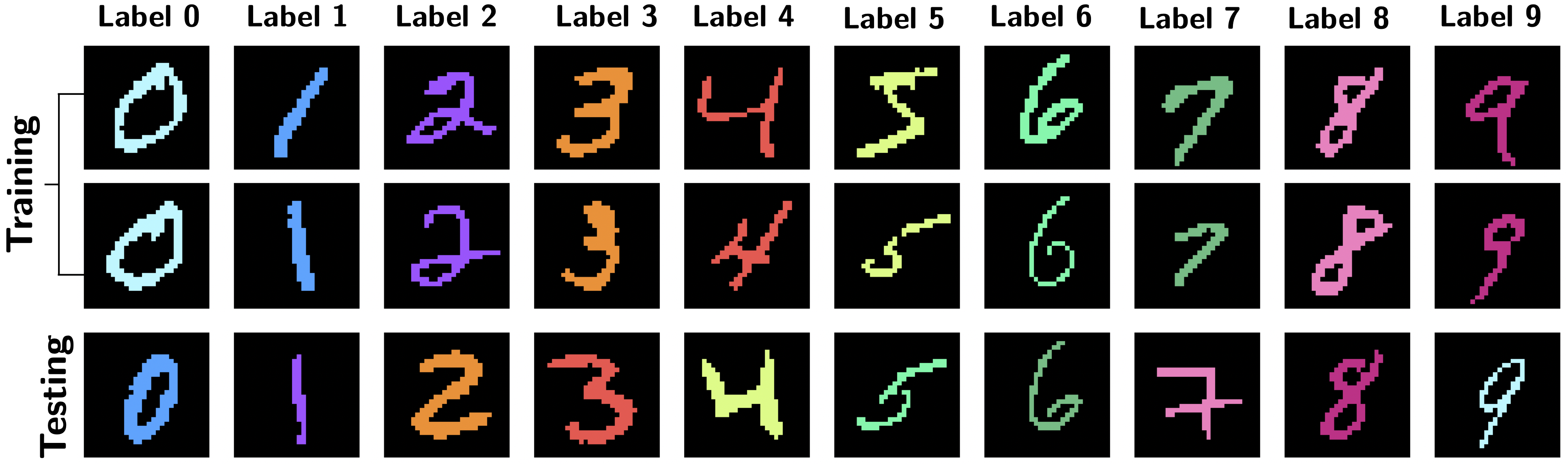}
    \caption{\small{\extendedcoloredmnist: Illustration of samples corresponding to a 10-digit classification task. During training, each class label is spuriously correlated with a unique color. While testing, this correlation is significantly reduced as each image is colored with the color corresponding to its succeeding label. }}
    \label{fig:multiclass_colored_mnist_dataset}
\end{figure}

 Further, since all the previous settings were binary classification tasks,  we extend the standard datasets \coloredmnist, \coloredfashionmnist and \cifar over multi class classification \cite{choe2020empirical}. Specifically, we extend the number of classes from 2 to 5 and 10. For \coloredfashionmnist and \coloredmnist,  a unique color is assigned to each output class such that the label is highly correlated ($\sim 80\% - 90\%$) with the color in the training set. In the test set, these correlations are significantly reduced ($10\%)$ by allowing high spurious correlations with the color of the following class. For instance, in testing, the color corresponding to class label 8 would be the one which was heavily corrected with class label 9 while training. This reduces the original spurious correlations and is hence useful for evaluating the extent to which the trained model has learned the invariant features. A sample batch of images elucidating the above construction is shown in Figure \ref{fig:multiclass_colored_mnist_dataset}.
We do not construct a multi-class classification setup for \cifar as it is difficult to find a unique spatial location in the image corresponding to each class (e.g. in case of 10-class classification).

\subsection{Experimental Setup}
\subsubsection{Architecture Details}
For all the approaches using a fixed representation i.e. \firm, \ffictplay, \textit{parallelized} \firm and \textit{parallelized} \ffictplay, we use the architecture mentioned below. 
The architecture used to train a predictor at each client is a multi-layered perceptron with three fully connected (FC) layers. The details of layers are as follows:
\begin{itemize}
    \item Flatten: A flatten layer that converts the input of shape \texttt{(Batch Size, Length, Width, Depth)} into a tensor of shape \texttt{(Batch Size, Length * Width * Depth)}
    \item FC1: A fully connected layer with an output dimension of 390, followed by \texttt{ELU} non-linear activation function
    \item FC2: A fully connected layer with an output dimension of 390, followed by \texttt{ELU} non-linear activation function
    \item FC3: A fully connected layer with an output dimension of 2 (Classification Layer)
\end{itemize}
This same architecture is used across approaches with fixed representation.
\par For the approaches with trainable representation i.e. \virm, \vfictplay, \textit{parallelized} \virm and \textit{parallelized} \vfictplay,  we use the following
architecture for the representation learner 
\begin{itemize}
    \item Flatten: A flatten layer that converts the input of shape \texttt{(Batch Size, Length, Width, Depth)} into a tensor of shape \texttt{(Batch Size, Length * Width * Depth)}
    \item FC1: A fully connected layer with an output dimension of 390, followed by \texttt{ELU} non-linear activation function
\end{itemize}

The output from FC1 is fed into the following architecture, which is used as the base network to train the predictor at each client:
\begin{itemize}
    \item FC1: A fully connected layer with an output dimension of 390, followed by \texttt{ELU} non-linear activation function
    \item FC2: A fully connected layer with an output dimension of 390, followed by \texttt{ELU} non-linear activation function
    \item FC3: A fully connected layer with an output dimension of 2 (Classification Layer)
\end{itemize}
This architecture is used across approaches with variable representation.

\subsubsection{Optimizer and other hyperparameters}
We use a different set of hyper-parameters based on the dataset. In particular, \\
\begin{itemize}
    \item \coloredmnist and \coloredfashionmnist: For fixed representation, we use Adams optimizer with a learning rate of 2.5e-4 across all experiments (sequential or parallel). For the variable representation, we use Adams optimizer with a learning rate of 2.5e-5 for the representation learner and the same optimizer with a learning rate of 2.5e-4 for the predictor at each client. 
    \item \cifar: For fixed representation, we use Adams optimizer with a learning rate of 1.0e-4 across all experiments (sequential or parallel). For the variable representation, we use Adams optimizer with a learning rate of 9.0e-4 for the representation learner and the same optimizer with a learning rate of 1.0e-4 for the predictor at each client.
\end{itemize}

For all the experiments, we fix the batch size to 256 and optimize the Cross Entropy Loss. We use the same termination criterion as in \cite{ahuja2020invariant}.  Specifically, we stop training when the observed oscillations become stable and the ensemble model is in a lower training accuracy state. We choose a training threshold and terminate the training as soon as the training accuracy drops below this value. In order to ensure stability of oscillations, we set a period of warm start. In this period, the training is not stopped even if the accuracy drops below the threshold. For variable representation, the duration of this warm start period is set to the number of training steps in an epoch i.e. (training data size/ batch size). However, for the approaches with fixed representation, this period is fixed to $N$ rounds where $N$ is the number of training clients. In particular, for two clients, the warm start period ends as soon as the second client finishes playing its optimal strategy for the first time. \\

\subsection{Additional Results and Analysis}
\subsubsection{Robustness to the number of Outcomes}
As described in Section \ref{multiclassdatasets}, we test the robustness of our approach to an increase in number of output classes. We compare \firm and \textit{parallelized} \firm across 2-digit, 5-digit and 10-digit classification for \coloredmnist and \coloredfashionmnist. 
\par As shown in Tables \ref{tab:multiple_outcomes_mnist} and \ref{tab:multiple_outcomes_fashionmnist}, both the sequential and the parallel version of \irmgames i.e. \firm and \textit{parallelized} \firm respectively are robust to an increase in the number of output classes. For both the datasets, \textit{parallelized} \firm performs at par or better than \firm.
\begin{table}[htb!]
\caption{\small{\coloredmnist:  Comparison of \firm and \firm (Parallel) with increasing number of output classes, in terms of the training and testing accuracy (mean $\pm$ std deviation). Here `Seq.' is an abbreviation used for `Sequential', which denotes \firm.}}
\centering
\begin{adjustbox}{width=0.65\textwidth}
\begin{tabular}{llll}
\toprule
 Type & \# Classes & \trainacc & \testacc  \\
\midrule
\multirow{3}{*}{\rotatebox[origin=c]{90}{Seq.}} &	2	& 75.13	$\pm$	1.38 &68.40	$\pm$ 1.83	\\
&	5	&	79.39	$\pm$	0.91	&	69.90	$\pm$	3.16	\\
&	10	&	82.27	$\pm$	0.76	&	69.22	$\pm$	3.11	\\
\midrule
\multirow{3}{*}{\rotatebox[origin=c]{90}{Parallel}} &	2	& 71.71 $\pm$ 8.23 &	69.73 $\pm$ 2.12 \\
&	5	&	78.61	$\pm$	2.86	&	68.42	$\pm$	2.54	\\
&	10	&	82.17	$\pm$	1.21	&	69.29	$\pm$	3.17	\\
\bottomrule
\end{tabular}
\end{adjustbox}
\label{tab:multiple_outcomes_mnist}
\end{table}

\begin{table}[htb!]
\caption{\small{\coloredfashionmnist: Comparison of \firm and \firm (Parallel) with increasing number of output classes, in terms of the training and testing accuracy (mean $\pm$ std deviation). Here `Seq.' is an abbreviation used for `Sequential', which denotes \firm.}}
\centering
\begin{adjustbox}{width=0.65\textwidth}
\begin{tabular}{llll}
\toprule
 Type & \# Classes & \trainacc & \testacc  \\
\midrule
\multirow{3}{*}{\rotatebox[origin=c]{90}{Seq.}} &	2	&	50.36 $\pm$	 2.78		&	47.36 	$\pm$	 4.33	\\
&	5	&	77.28	$\pm$	1.54	&	69.35	$\pm$	0.66	\\
&	10	&	80.02	$\pm$	0.38	&	71.22	$\pm$	3.00	\\
\midrule
\multirow{3}{*}{\rotatebox[origin=c]{90}{Parallel}} &	2	&		55.06 $\pm$ 2.04		&	52.07 $\pm$ 1.60		\\
&	5	&	77.61	$\pm$	1.51	&	70.83	$\pm$	0.96	\\
&	10	&	80.12	$\pm$	0.78	&	70.39	$\pm$	2.28	\\
\bottomrule
\end{tabular}
\end{adjustbox}
\label{tab:multiple_outcomes_fashionmnist}
\end{table}

\FloatBarrier
\subsection{Effect of Simultaneous BRD}
\begin{figure}[htb!]
\centering
  \subfloat[]{\includegraphics[width=0.45\textwidth]{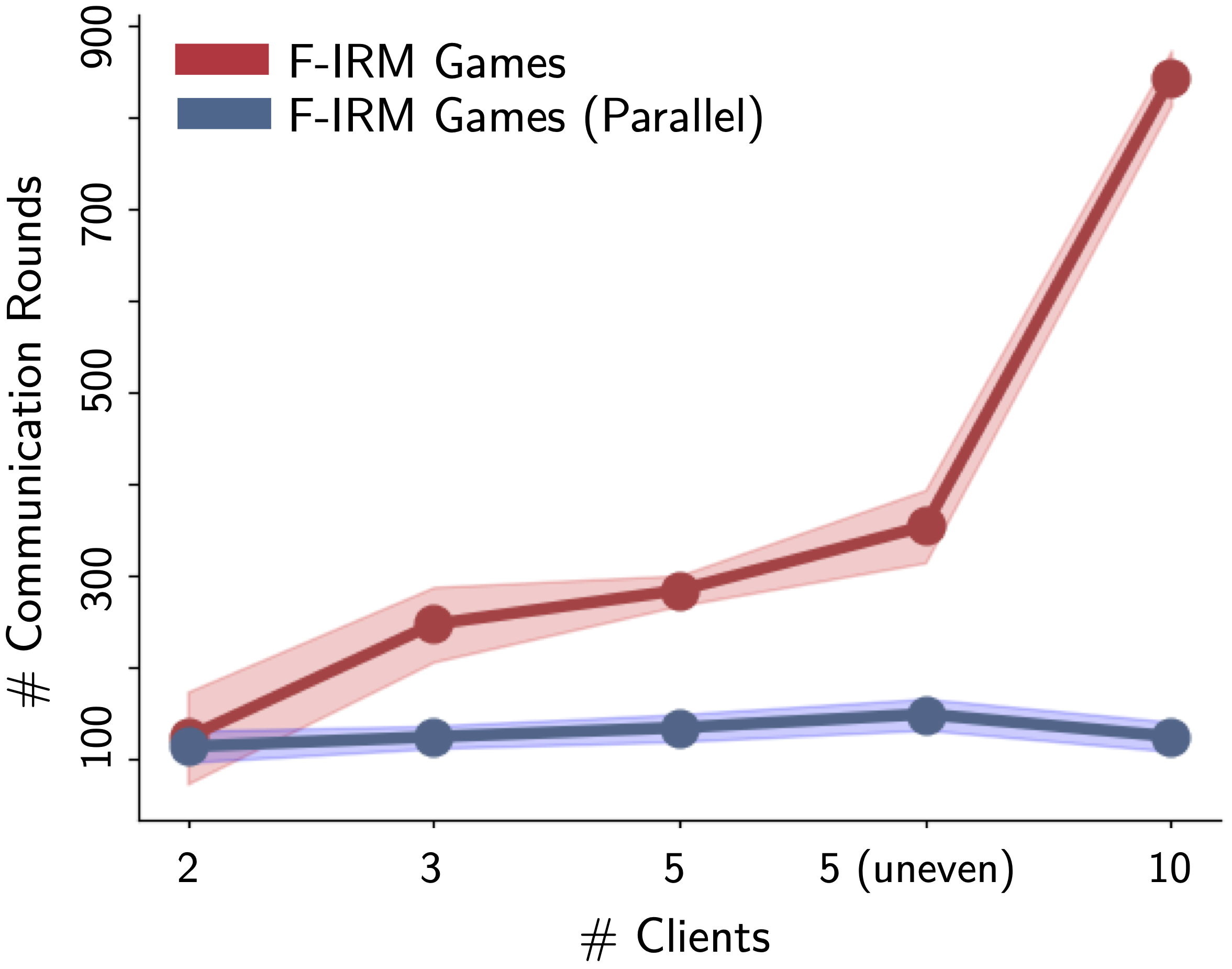}\label{fig:sequential_vs_parallel_ablation_b}}
  \hfill
  \subfloat[]{\begin{adjustbox}{width=0.5\textwidth}
\begin{tabular}{llll}
\toprule
 Type & \# clients & \trainacc & \testacc  \\
\midrule
\multirow{5}{*}{\rotatebox[origin=c]{90}{Sequential}} &	2	&	59.85	$\pm$	7.80	&	65.99	$\pm$	2.44	\\
&	3	&	53.76	$\pm$	5.46	&	67.22	$\pm$	0.93	\\
&	5	&	57.64	$\pm$	2.57	&	67.75	$\pm$	0.52	\\
&	5 (uneven)	&	56.48	$\pm$	2.10	&	67.17	$\pm$	0.84	\\
&	10	&	57.40	$\pm$	1.67	&	68.94	$\pm$	1.35	\\
\midrule
\multirow{5}{*}{\rotatebox[origin=c]{90}{Parallel}}   &	2	&	58.29	$\pm$	4.31	&	69.38	$\pm$	3.61	\\
&	3	&	61.10	$\pm$	1.68	&	69.88	$\pm$	2.82	\\
&	5	&	60.00	$\pm$	5.78	&	68.57	$\pm$	2.86	\\
&	5 (uneven)	&	66.82	$\pm$	2.55	&	66.95	$\pm$	2.41	\\
&	10	&	63.86	$\pm$	3.60	&	70.71	$\pm$	2.99	\\
\bottomrule
\end{tabular}
\end{adjustbox}}
\caption{\small{\coloredfashionmnist: (a) Number of communication rounds required to achieve the Nash equilibrium versus the number of clients in the FL setup; (b) Comparison of \firm and \firm (Parallel) with an increase in the number of clients, in terms of training and testing accuracy (mean $\pm$ std deviation).}}
\label{fig:parallel_vs_sequential_fashion_mnist}
\end{figure}

\begin{figure}[htb!]
\centering
  \subfloat[]{\includegraphics[width=0.45\textwidth]{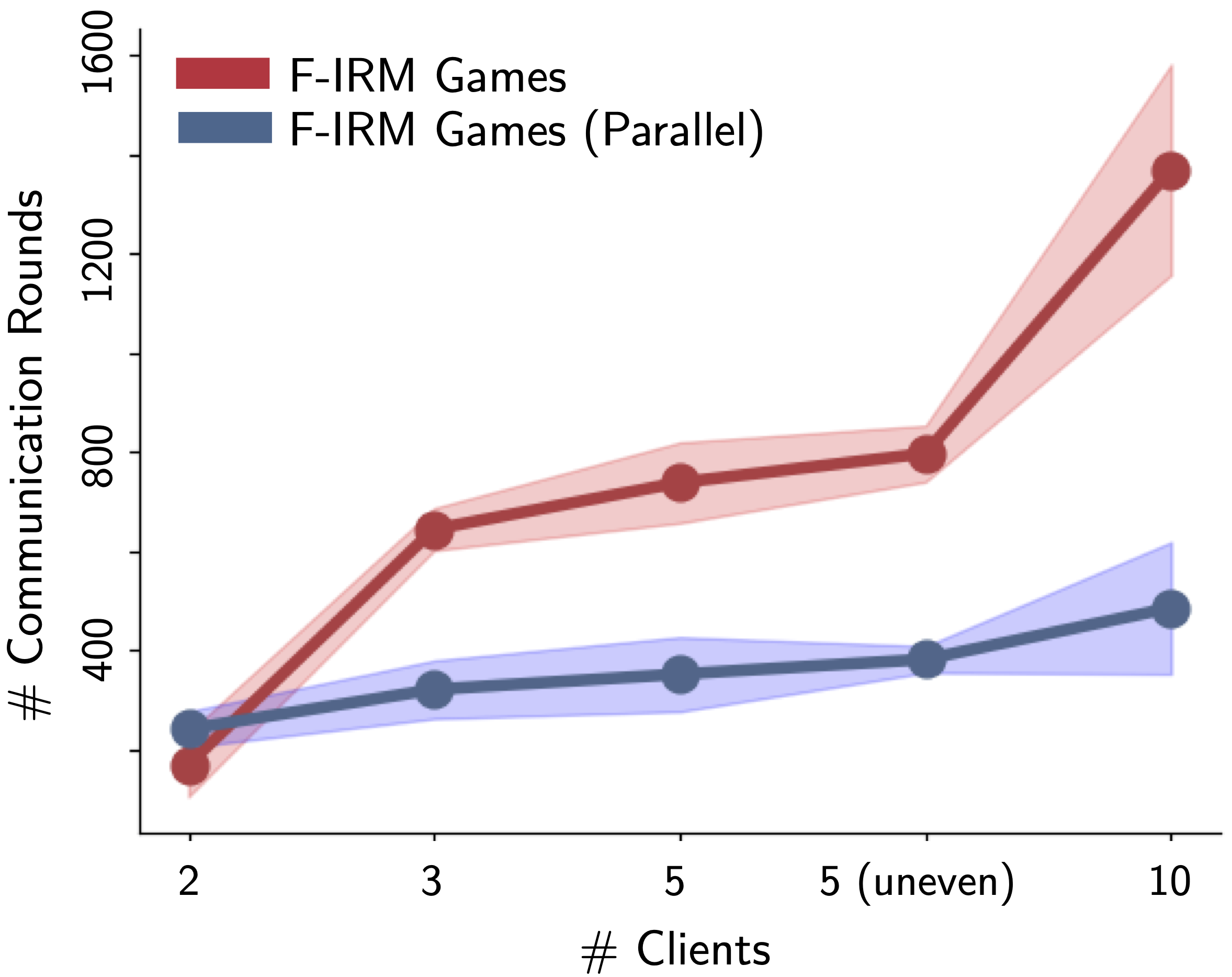}\label{fig:sequential_vs_parallel_ablation_a}}
  \hfill
  \subfloat[]{\begin{adjustbox}{width=0.50\textwidth}
\begin{tabular}{llll}
\toprule
 Type & \# clients & \trainacc & \testacc  \\
\midrule
\multirow{5}{*}{\rotatebox[origin=c]{90}{Sequential}} &	2	&	50.37	$\pm$	3.07	&	48.70	$\pm$	4.12	\\
&	3	&	48.61	$\pm$	1.40	&	59.31	$\pm$	12.45	\\
&	5	&	56.70	$\pm$	6.25	&	45.27	$\pm$	0.42	\\
&	5 (uneven)	&	62.62	$\pm$	6.84	&	47.27	$\pm$	3.81	\\
&	10	&	52.40	$\pm$	1.44	&	50.97	$\pm$	0.92	\\
\midrule
\multirow{5}{*}{\rotatebox[origin=c]{90}{Parallel}}  &	2	&	57.13	$\pm$	4.82	&	50.50	$\pm$	2.70	\\
&	3	&	55.15	$\pm$	2.99	&	53.97	$\pm$	1.06	\\
&	5	&	54.72	$\pm$	1.07	&	51.91	$\pm$	0.50	\\
&	5 (uneven)	&	54.77	$\pm$	2.39	&	53.40	$\pm$	1.83	\\
&	10	&	52.67	$\pm$	1.11	&	53.50	$\pm$	0.84	\\
\bottomrule
\end{tabular}
\end{adjustbox}}
\caption{\small{\cifar: (a) Number of communication rounds required to achieve the Nash equilibrium versus the number of clients in the FL setup; (b) Comparison of \firm and \firm (Parallel) with an increase in the number of clients, in terms of training and testing accuracy (mean $\pm$ std deviation).}}
\label{fig:parallel_vs_sequential_cifar}
\end{figure}

Similar to the experiments conducted for \coloredmnist, where we compared \firm and \textit{parallelized} \firm across an increase in the number of clients, we replicate the same setup for \coloredfashionmnist and \cifar. We report the results on both datasets in Figures \ref{fig:parallel_vs_sequential_fashion_mnist} and \ref{fig:parallel_vs_sequential_cifar}. Consistent with the results on \coloredmnist, as the number of clients in the FL system increases, there is a sharp increase in the number of communication rounds required to reach equilibrium. However, the same doesn’t hold true for \textit{parallelized} \firm. Further, the accuracy achieved by \textit{parallelized} \firm is comparable or better than that achieved by \firm. 

\FloatBarrier
\subsubsection{Effect of Memory Ensemble}
\begin{figure}[htb!]
    \centering
\includegraphics[width=0.55\linewidth]{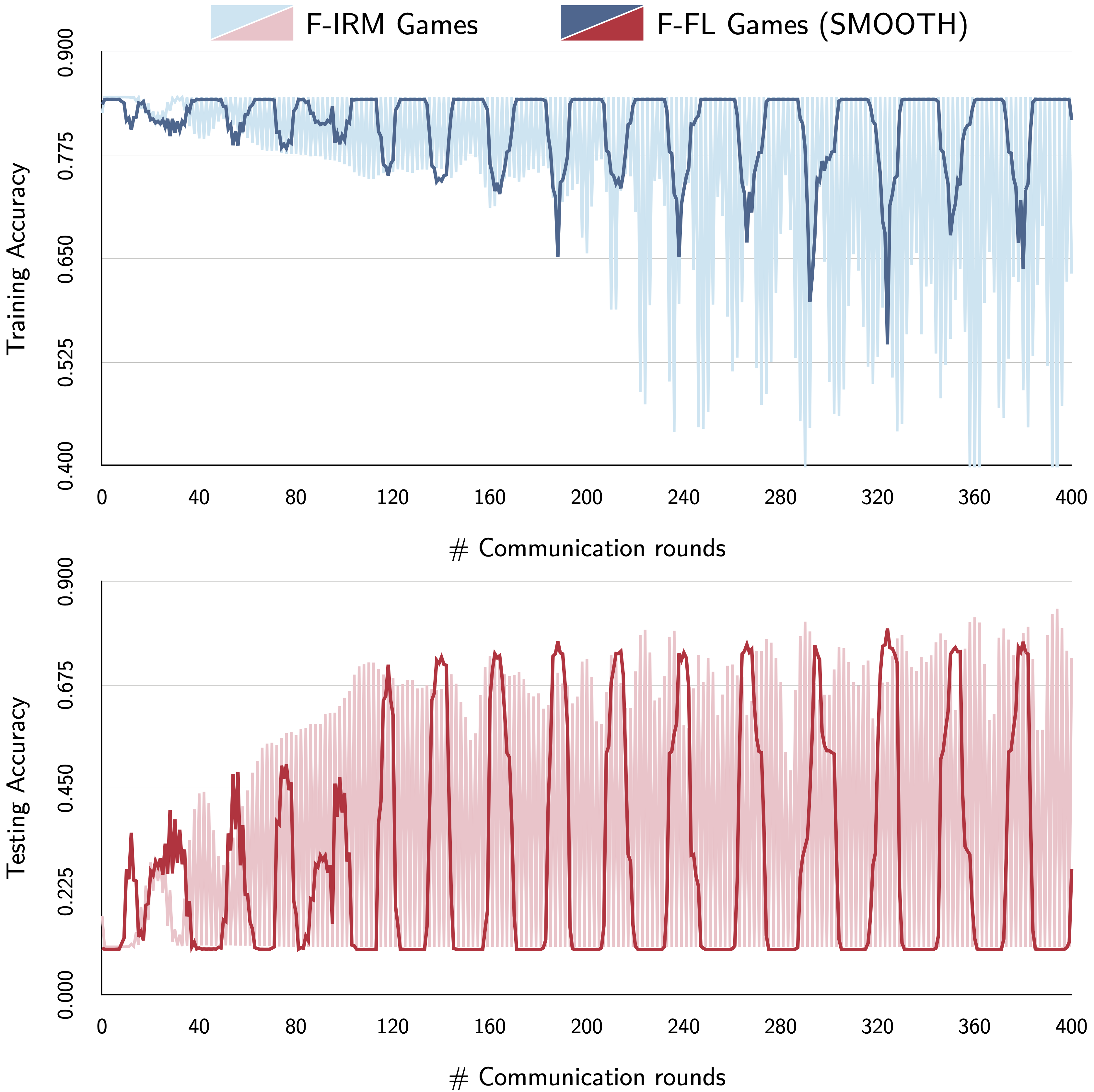}
    \caption{\small{\coloredfashionmnist: Evolution of Training and Testing Training for \firm and \ffictplay using a buffer size of 5, over the number of communication rounds}}
    \label{fig:fictplay_vs_standard_fashionmnist}
\end{figure}

\begin{figure}[htb!]
    \centering
\includegraphics[width=0.55\linewidth]{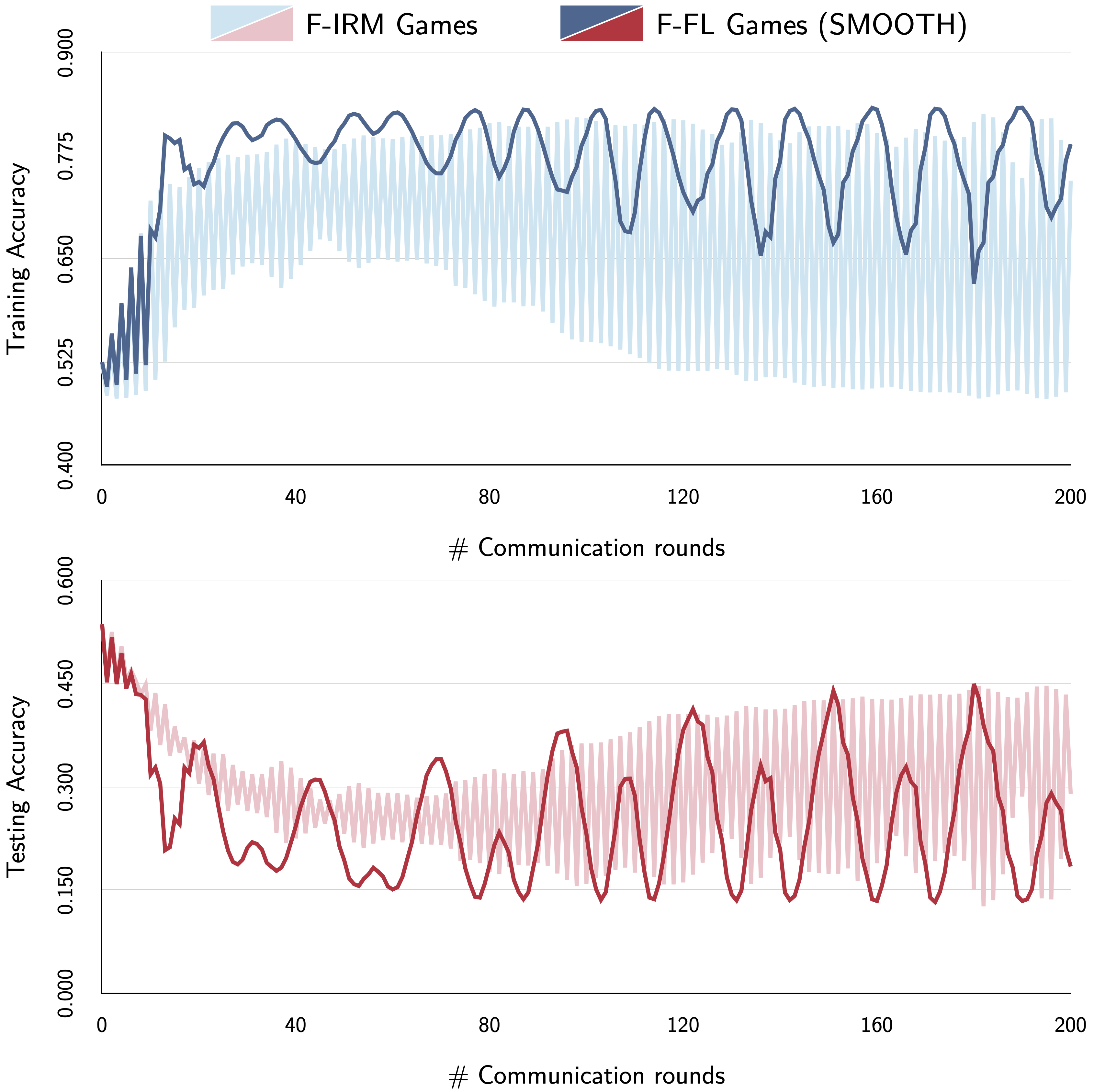}
    \caption{\small{\cifar: Evolution of Training and Testing Training for \firm and \ffictplay using a buffer size of 5, over the number of communication rounds}}
    \label{fig:fictplay_vs_standard_cifar}
\end{figure}

Similar to the experiments conducted for \coloredmnist, where we compared \firm and \ffictplay, we replicate the same setup for \coloredfashionmnist and \cifar. We report the results on both datasets in Figures \ref{fig:fictplay_vs_standard_fashionmnist} and \ref{fig:fictplay_vs_standard_cifar}.

Consistent with the results on \coloredmnist, performance curves oscillate at each step for \firm while the oscillations in \ffictplay are observed after an interval of roughly 40 rounds for both the datasets. Further, \ffictplay also achieves high testing accuracy. This implies that it does not rely on the spurious features to make predictions. Similar performance evolution curves are also observed for \textit{parallelized} \ffictplay with an added benefit of faster convergence as compared to \firm and \ffictplay.

\subsubsection{Effect of using Gradient Descent (GD) for $\phi$}
\begin{figure}[htb!]
    \centering
\includegraphics[width=0.55\linewidth]{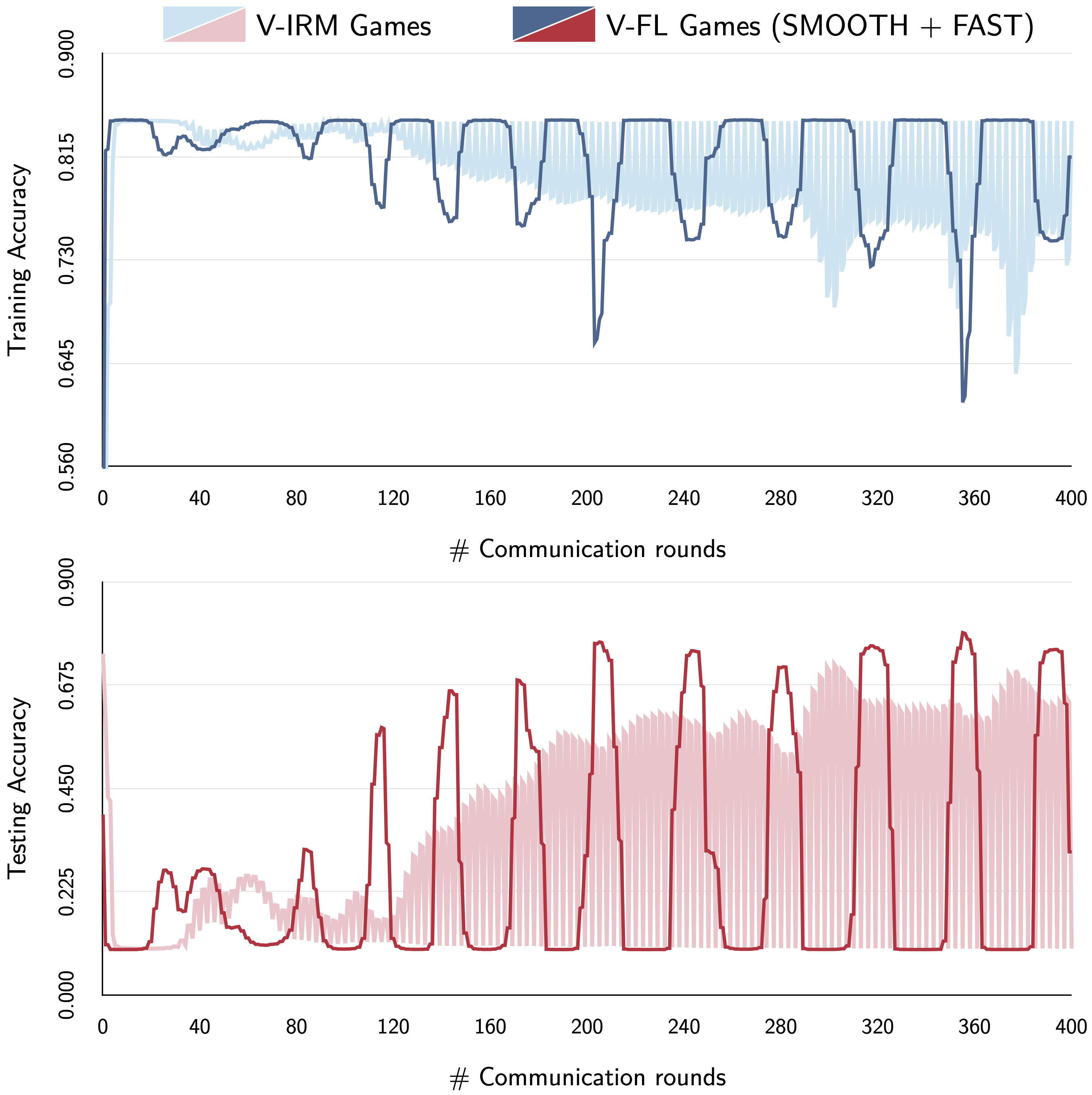}
    \caption{\small{\coloredfashionmnist: Evolution of Training and Testing Training for \virm and \finalalgo using a buffer size of 5, over the number of communication rounds
    }}
    \label{fig:vfastfictplay_vs_vstandard_fashionmnist}
\end{figure}

\begin{figure}[htb!]
    \centering
\includegraphics[width=0.55\linewidth]{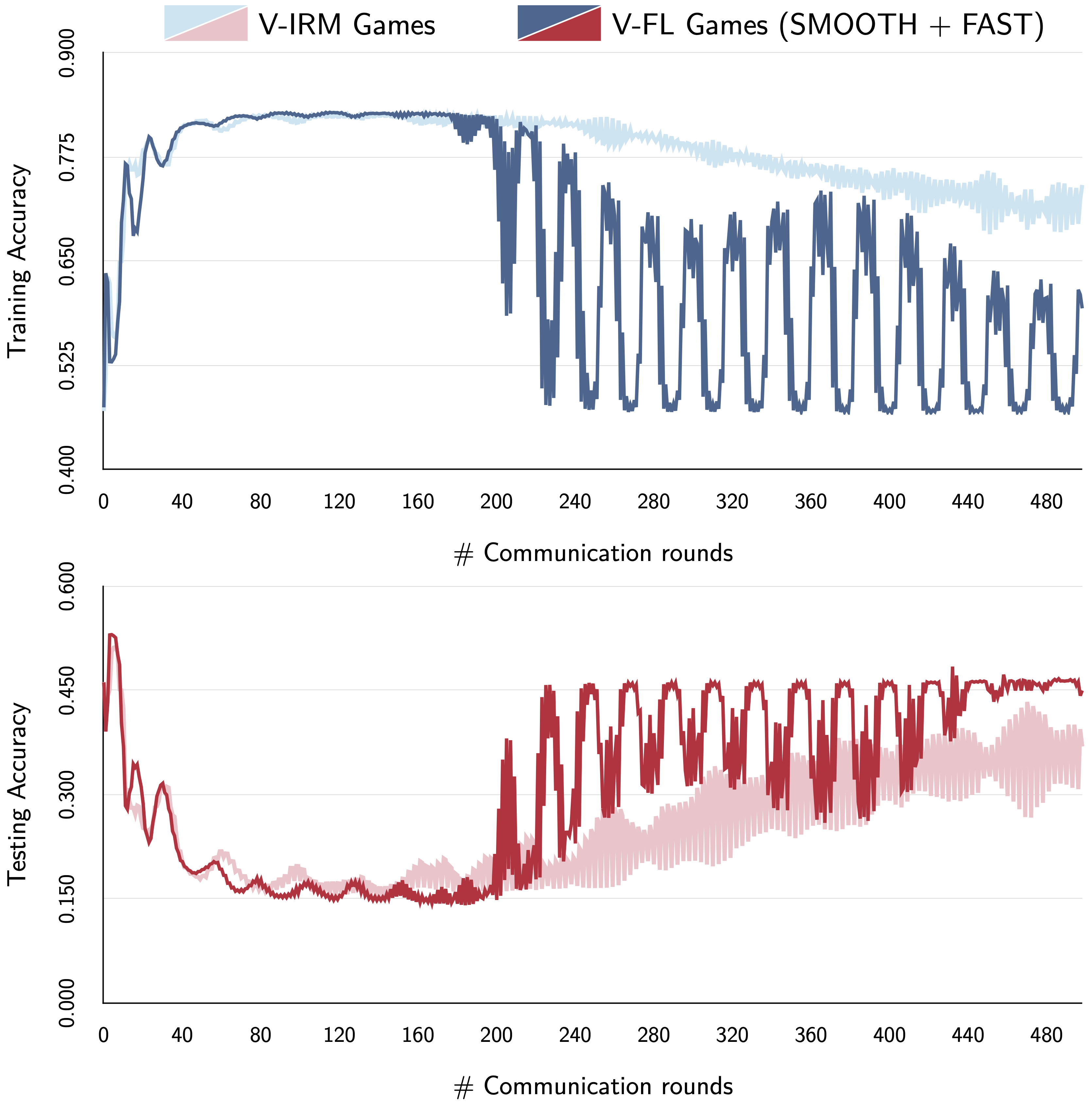}
    \caption{\small{\cifar: Evolution of Training and Testing Training for \virm and \finalalgo using a buffer size of 5, over the number of communication rounds}}
    \label{fig:vfastfictplay_vs_vstandard_cifar}
\end{figure}

Similar to the experiments conducted for \coloredmnist, where we compared \virm and \finalalgo, we replicate the same setup for \coloredfashionmnist and \cifar. We report the results on both datasets in Figures \ref{fig:vfastfictplay_vs_vstandard_fashionmnist} and \ref{fig:vfastfictplay_vs_vstandard_cifar}. Consistent with the results on \coloredmnist, \finalalgo is able to achieve significantly higher testing accuracy in fewer communication rounds as compared to \virm on both the datasets.

\FloatBarrier
\subsubsection{Effect of exact best response }
Similar to the experiments conducted for \coloredmnist, where we studied the influence of increasing the amount of local computation at each client,  we replicate the setup for \coloredfashionmnist and \cifar. 

\begin{figure}[htb!]
    \centering
\includegraphics[width=0.55\linewidth]{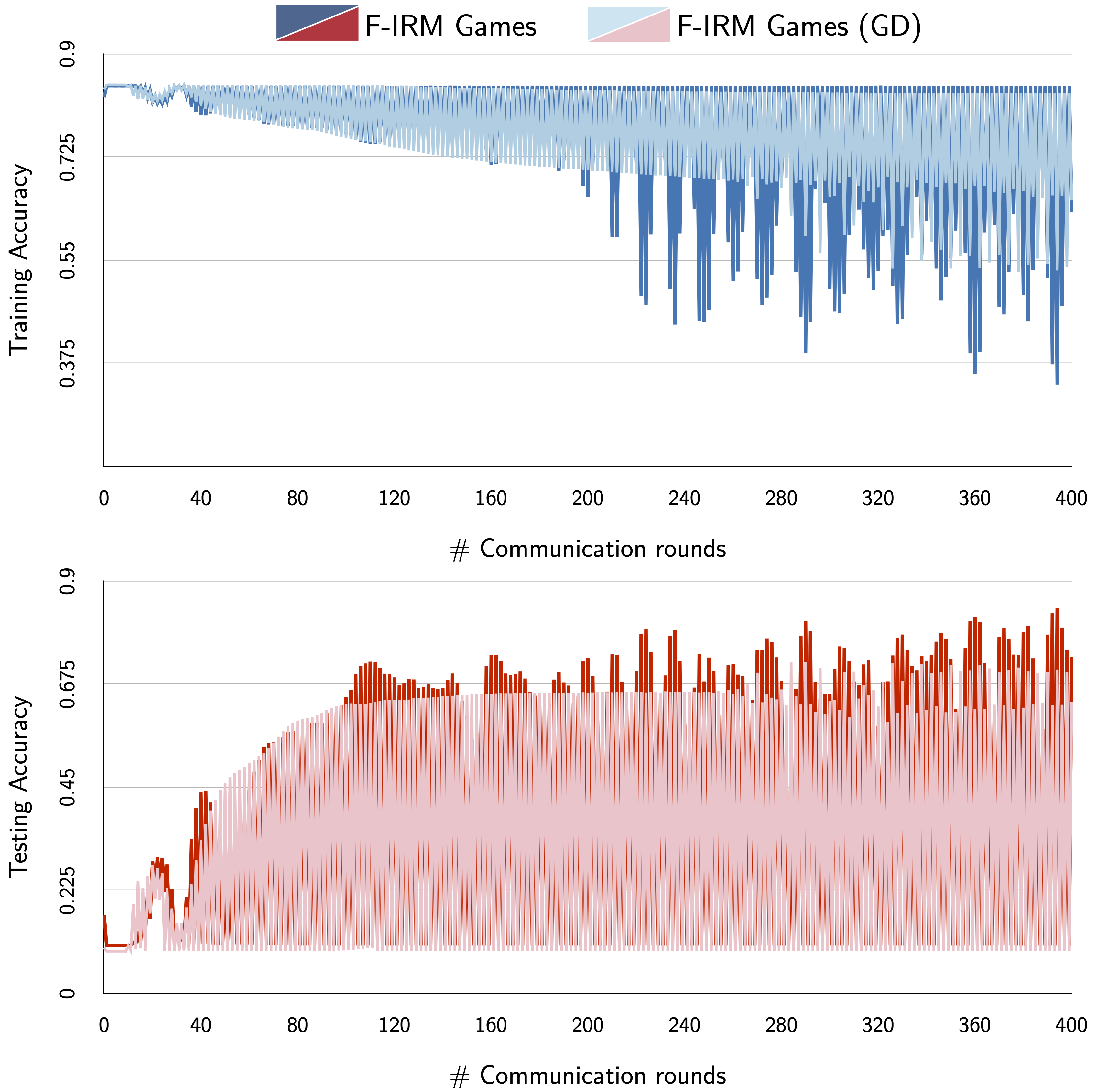}
    \caption{\small{\coloredfashionmnist: Evolution of Training and Testing Training when doing a gradient descent to update the predictor at each client versus the standard training paradigm i.e. \firm, over the number of communication rounds.
    }}
    \label{fig:local_compute_vs_standard_fashionmnist}
\end{figure}

\begin{figure}[htb!]
    \centering
\includegraphics[width=0.55\linewidth]{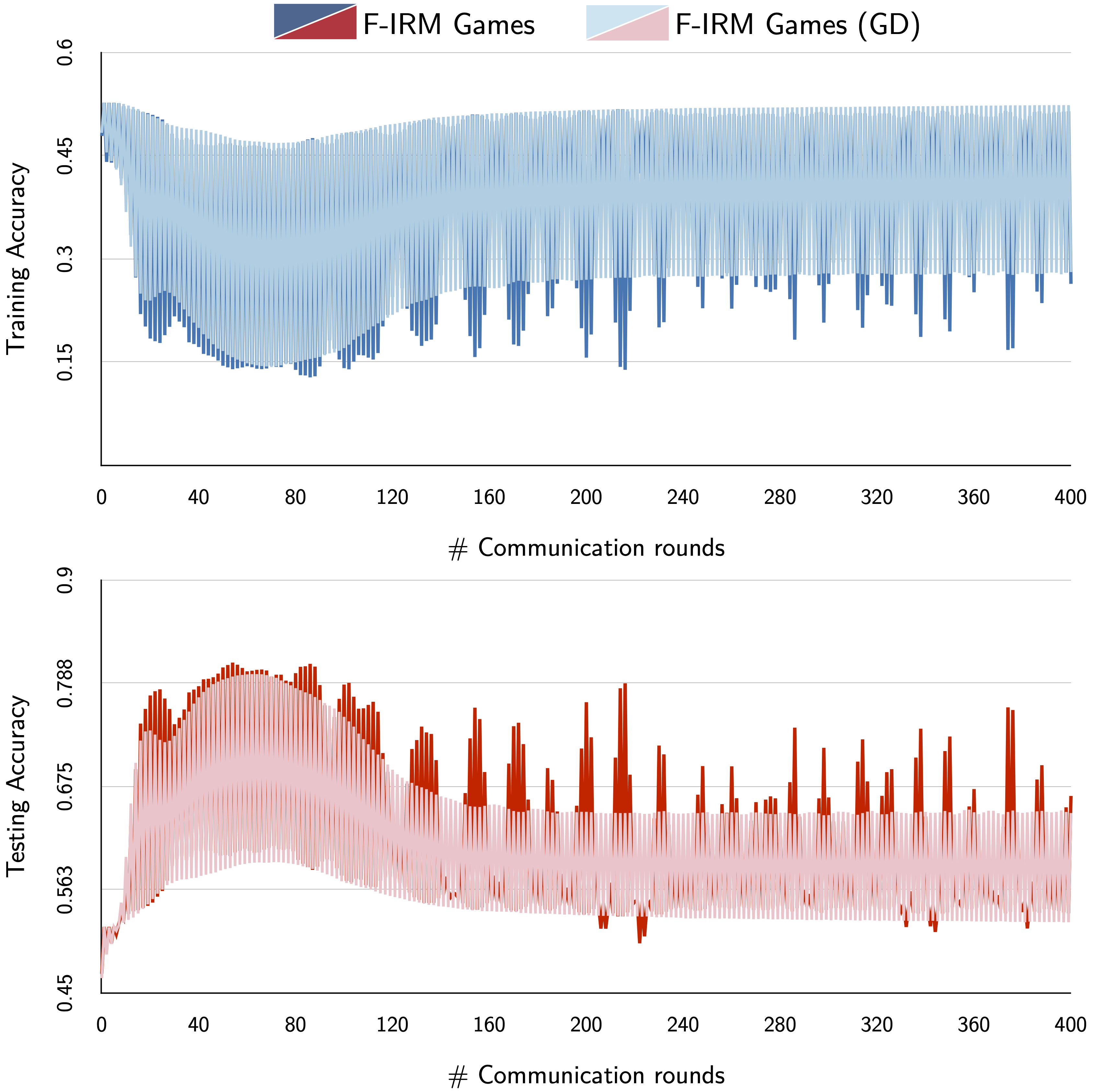}
    \caption{\small{\cifar: Evolution of Training and Testing Training when doing a gradient descent to update the predictor at each client versus the standard training paradigm i.e. \firm, over the number of communication rounds.}}
    \label{fig:local_compute_vs_standard_cifar}
\end{figure}

Consistent with the results on \coloredmnist and as shown in Figures \ref{fig:local_compute_vs_standard_fashionmnist} and \ref{fig:local_compute_vs_standard_cifar}, even when the number of local steps at each client reaches its maximum i.e. (training data size/ mini-batch size)), the trained models are able to achieve high testing accuracy at equilibrium. Further, we can also conclude from Figures \ref{fig:local_compute_vs_standard_fashionmnist} and \ref{fig:local_compute_vs_standard_cifar} that this setup achieves faster convergence with a slight change in the training accuracy at equilibrium.
As observed from the empirical results and discussed by \cite{ahuja2021linear}, \irmgames exhibits convergence guarantees despite an increase in the amount of local compute. This has direct implications towards training stability and building communication efficient FL systems.


\clearpage
\end{document}